\documentclass[10pt,twocolumn,letterpaper]{article}
\usepackage[T1]{fontenc} 

\usepackage{cvpr}              

\newcommand{\Dir}{\mathrm{Dir.}}
\newcommand{\Der}{\mathrm{Der.}}
\newcommand{\Inv}{\mathrm{Inv.}}
\newcommand{\Fid}{F}
\newcommand{\Smo}{M}
\newcommand{\GTI}{G}
\newcommand{\src}{\mathbf{S}}
\newcommand{\gt}{\mathbf{G}}
\newcommand{\cand}{\mathbf{C}}
\newcommand{\gallerycaptionfont}{\footnotesize}

\definecolor{cvprblue}{rgb}{0.21,0.49,0.74}
\usepackage[pagebackref,breaklinks,colorlinks,allcolors=cvprblue]{hyperref}
\def\confName{CVPR}
\def\confYear{2027}

\newif\ifwithsupp
\withsupptrue

\title{VEDIT-Bench: A Unified Benchmark for Evaluating Coding Agents on Programmatic Video Editing from Undescribed Instructions}

\author{
Haoyu Yao$^{1}$ \quad
Chenguo Lin$^{1}$\thanks{Project lead.} \quad
Yike Xiang$^{2}$ \quad
Ruize Du$^{1}$ \quad
Fengyuan Ran$^{1}$ \quad \\
Qingmiao Hu$^{3}$ \quad
Xuesong Duan$^{4}$ \quad
Yadong Mu$^{1}$\thanks{Corresponding author.} \quad\\
$^1$ Peking University \quad
$^2$ Wuhan University\\
$^3$ Nanyang Technological University \quad
$^4$ Paris-Saclay University
}

\begin{document}
\setcounter{footnote}{1}
\maketitle
\begin{abstract}
Programmatic dynamic visual artifacts, such as animations, simulations, and interactive visualizations, are specified by executable programs, which makes them parametric, controllable, and reproducible. In practice, coding agents are often tasked with editing these artifacts given only a rendered video and an underspecified natural language instruction that states the primary modification while leaving its structural, quantitative, and temporal consequences implicit. We formulate this task as video-conditioned programmatic editing from undescribed instructions: without access to the original source code, an agent must reconstruct the underlying scene, execute the explicit edit, propagate its implicit consequences over time, and preserve all unaffected content and dynamics. To evaluate this capability, we introduce VEDIT-Bench, which comprises 678 editing tasks across 226 source scenes spanning seven domains and three program families. These tasks are organized into three escalating difficulty tiers, namely L1 for local edits, L2 for linked updates, and L3 for rule or structural changes, and are paired with time-anchored Direct, Derived, and Invariant metrics. Evaluating eight frontier coding agents reveals that while local edits barely differentiate models, performance degrades sharply on L3 tasks, accompanied by a substantial rank reshuffle and a pronounced drop in dependent updates. These results indicate that current coding agents can execute the stated change and preserve the rest of the video, but struggle to infer the latent intent of an instruction and carry its consequences into the program. The project page is \href{https://haoyuyao.github.io/VEDIT-Bench}{https://haoyuyao.github.io/VEDIT-Bench}.
\end{abstract}
    
\section{Introduction}
\label{sec:intro}

\begin{figure*}[t]
\centering
\includegraphics[width=\linewidth]{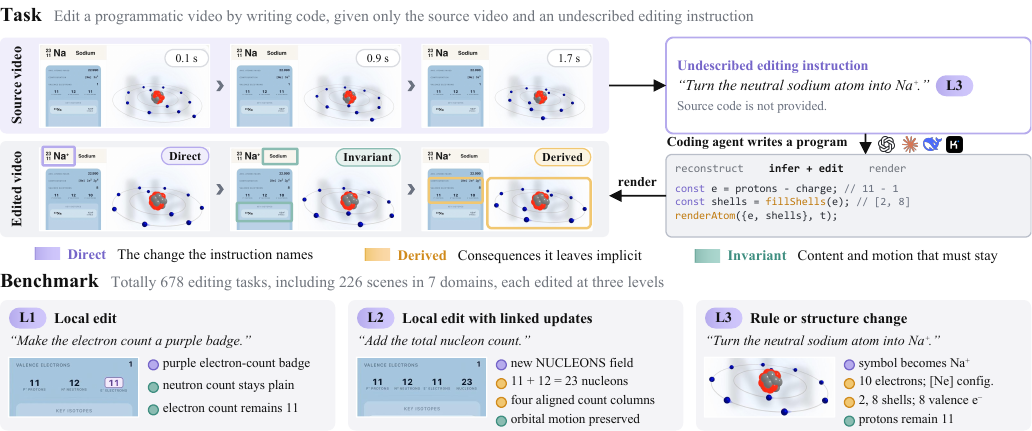}
\caption{\textbf{Overview of VEDIT-Bench.} Given only a source video and an undescribed instruction with hidden intent, agents reconstruct and edit the video via code. \textbf{Top:} Changing Na to Na$^{+}$ requires updating the symbol (\emph{Direct}) and implicitly adjusting electron count (11$\rightarrow$10), configuration, shells, and valence (\emph{Derived}), while preserving 11 protons (\emph{Invariant}). \textbf{Bottom:} Difficulty levels illustrated on the same scene: local appearance (L1), linked quantity addition (L2), and reasoning structural change (L3).}
\label{fig:task}
\end{figure*}

Programmatic dynamic visual artifacts, \ie, animations, simulations, and interactive visualizations whose visual elements, scene structure, and temporal behavior are specified by executable programs, are ubiquitous in scientific visualization, education, physical simulation, and data presentation. Unlike raw videos, which store content as pixels, such artifacts are parametric, controllable, and reproducible: changing a single parameter consistently updates every element that depends on it. With the rapid progress of multimodal large language models (MLLMs) and coding agents~\cite{sweagent,openhands}, recovering executable programs from visual references and faithfully reproducing dynamic content has become increasingly feasible~\cite{animation2code,4dcodebench}.

However, reconstructing these artifacts is often only an intermediate step, as agents typically need to edit existing artifacts using partial natural-language instructions. Such instructions describe only the primary modification and leave the resulting changes to related elements, quantities, and temporal behaviors implicit. Correctly executing an edit therefore requires a coding agent to infer the user's underlying intent and reason about the structural, quantitative, and temporal dependencies that are not explicitly stated. For example, given an unchanged resistive load, increasing a circuit's supply voltage from 5\,V to 10\,V requires not only updating the voltage but also doubling the current and quadrupling the power dissipated by that load, as well as adjusting the corresponding visual dynamics. This challenge is further compounded when the original source code is unavailable and only a rendered video is provided. The agent must then reconstruct the underlying scene structure and dynamics from pixels, infer the full scope of the intended edit, and generate a revised executable program that incorporates all necessary changes while preserving unaffected content, as shown in Figure~\ref{fig:task}. This motivates the central question of our work: \emph{Given a video and an undescribed editing instruction, can coding agents infer the intended edits and consequences, then generate executable code that produces a consistent video while preserving unaffected content?}

Existing benchmarks address only parts of this problem, as shown in Table~\ref{tab:compare}. Instruction-guided video editing benchmarks~\cite{five_bench,ivebench,omniedit_bench} evaluate instruction following, content preservation, and temporal consistency, but focus on perceptual plausibility rather than quantitative and logical correctness (\eg{}, Ohm's law). Code-driven video generation and reconstruction benchmarks~\cite{vibevideobench,animation2code,4dcodebench,code2video_compass} evaluate executable video generation or the recovery of appearance and dynamics, but not editing. Code-based editing of static visuals such as charts and vector graphics~\cite{chartedit,svgeditbench} does not capture temporal dependencies. Consequently, no existing benchmark jointly evaluates video reconstruction, explicit editing, implicit dependency propagation, and preservation of unaffected content. Moreover, holistic video similarity is insufficient for this task, as small but important derived changes (\eg{}, a numerical label) may affect few pixels.

To bridge this gap, we introduce \textbf{VEDIT-Bench}, a unified benchmark designed to evaluate coding agents on video-conditioned programmatic editing with dependency inference and temporal reasoning. Given only a rendered source video and an underspecified natural-language instruction while the original source code remains withheld, an agent must synthesize a self-contained executable program that applies the explicit edit, propagates its implicit structural, quantitative, and temporal consequences, and preserves all unaffected scene content and dynamics. VEDIT-Bench comprises 678 editing tasks derived from 226 deterministic source scenes spanning seven domains, namely algorithm animations, charts, chemistry, circuits, mathematics, physics, and web interfaces, across animated SVG, HTML/JavaScript, and Python programs. Each source scene is instantiated across three escalating difficulty tiers: L1 for local edits, L2 for linked updates, and L3 for governing rule or structural changes. To overcome the limitations of holistic video similarity, we further propose time-anchored \textbf{Direct}, \textbf{Derived}, and \textbf{Invariant} semantic criteria paired with complementary pixel-level metrics to disentangle explicit edit execution, downstream causal propagation, and background preservation.

In total, we evaluate eight frontier coding agents across all three editing levels. Our empirical findings demonstrate that current agents still struggle to infer latent editing intent and translate its downstream consequences into programmatic video modifications. Specifically, while local edits at L1 barely differentiate top-performing systems, every evaluated agent suffers a sharp performance degradation when transitioning to L3 tasks involving rule or structural changes, accompanied by a substantial leaderboard reshuffle in which the second-ranked L1 system drops to seventh place. From L2 to L3, the mean \textbf{Derived} score falls by approximately 20 points compared to a 4.5-point drop in \textbf{Direct}, whereas visual fidelity outside the edited region remains largely unchanged. These results show that even when agents successfully implement the explicitly stated modification and preserve visual appearance, they frequently fail to propagate implicit dependencies across coupled elements and temporal events.

Our main contributions are summarized as follows:
\begin{itemize}
    \item We formalize \emph{video-conditioned programmatic editing with dependency inference and temporal reasoning}, requiring coding agents to reconstruct dynamic artifacts from raw videos, infer implicit editing intent, and synthesize executable programs without source code.
    \item We introduce VEDIT-Bench, comprising 678 video editing tasks across 226 scenes, seven domains, three program families, and three escalating difficulty levels from L1 to L3, paired with time-anchored \emph{Direct}, \emph{Derived}, and \emph{Invariant} metrics alongside pixel-level measures.
    \item We benchmark eight frontier coding agents across matched scenes and show that current models remain severely limited in inferring unstated editing intent and propagating dependent updates into executable video programs, particularly under structural modifications.
\end{itemize}

\section{Related Work}
\label{sec:related_work}

\begin{table}[t]
\centering
\caption{Comparison with related benchmarks. \emph{Dyn.}: the artifact changes over time. \emph{Code}: the output is an executable program. \emph{Edit}: the task modifies an existing artifact. \emph{Hidden}: the source program is not given. \emph{Derived}: consequences that the instruction leaves implicit are scored separately. A dash means not applicable.}
\label{tab:compare}
\small
\setlength{\tabcolsep}{4pt}
\begin{tabular}{@{}lccccc@{}}
\toprule
Benchmark & Dyn. & Code & Edit & Hidden & Derived \\
\midrule
\multicolumn{6}{@{}l@{}}{\textit{Instruction-guided Video Editing}} \\
FiVE-Bench~\cite{five_bench} & $\checkmark$ & $\times$ & $\checkmark$ & -- & $\times$ \\
IVEBench~\cite{ivebench} & $\checkmark$ & $\times$ & $\checkmark$ & -- & $\times$ \\
OmniEdit-Bench~\cite{omniedit_bench} & $\checkmark$ & $\times$ & $\checkmark$ & -- & $\times$ \\
\midrule
\multicolumn{6}{@{}l@{}}{\textit{Programmatic Video Generation}} \\
VibeVideoBench~\cite{vibevideobench} & $\checkmark$ & $\checkmark$ & $\times$ & -- & $\times$ \\
Animation2Code~\cite{animation2code} & $\checkmark$ & $\checkmark$ & $\times$ & $\checkmark$ & $\times$ \\
4DCodeBench~\cite{4dcodebench} & $\checkmark$ & $\checkmark$ & $\times$ & $\checkmark$ & $\times$ \\
Code2Video-Compass~\cite{code2video_compass} & $\checkmark$ & $\checkmark$ & $\times$ & -- & $\times$ \\
\midrule
\multicolumn{6}{@{}l@{}}{\textit{Programmatic Artifacts Editing}} \\
ChartEdit~\cite{chartedit} & $\times$ & $\checkmark$ & $\checkmark$ & $\checkmark$ & $\times$ \\
SVGEditBench~\cite{svgeditbench} & $\times$ & $\checkmark$ & $\checkmark$ & $\times$ & $\times$ \\
VAO-Bench~\cite{guo2026vaobench} & $\times$ & $\checkmark$ & $\checkmark$ & $\times$ & $\checkmark$ \\
\midrule
\textbf{VEDIT-Bench} (ours) & $\checkmark$ & $\checkmark$ & $\checkmark$ & $\checkmark$ & $\checkmark$ \\
\bottomrule
\end{tabular}
\end{table}

\noindent\textbf{Instruction-guided Video Editing Benchmarks.}
Text-driven video editing has evolved from per-video fine-tuning and attention-manipulation approaches~\cite{tuneavideo,fatezero,tokenflow} to instruction-based editors trained on synthetic paired data~\cite{insv2v} and training-free frameworks that propagate image edits through video generative models~\cite{anyv2v}. Benchmarks such as FiVE-Bench~\cite{five_bench}, IVEBench~\cite{ivebench}, and OmniEdit-Bench~\cite{omniedit_bench} assess instruction following, content preservation, and temporal consistency through embedding similarity, perceptual metrics, and MLLM-based judges. Since these methods operate in pixel or latent space, the benchmarks focus on whether the edited video \emph{looks} plausible. In contrast, the artifacts we study are governed by explicit rules and quantities, and an edit is correct only if all dependent values, elements, and events change consistently. VEDIT-Bench therefore requires executable programs and evaluates edits by their quantitative and logical correctness.

\noindent\textbf{Programmatic Video Generation Benchmarks.}
Prior work on multimodal code generation has largely focused on converting static visual inputs into executable code, spanning webpages~\cite{design2code,interaction2code}, scientific charts~\cite{chartmimic,plot2code}, and vector graphics~\cite{starvector,vgbench}. More recently, this paradigm has expanded to dynamic visual content: agentic pipelines synthesize Manim-based explanatory videos from text~\cite{theoremexplainagent,code2video}, while benchmarks such as VibeVideoBench~\cite{vibevideobench}, Animation2Code~\cite{animation2code}, 4DCodeBench~\cite{4dcodebench}, and Code2Video-Compass~\cite{code2video_compass} evaluate programmatic video synthesis or appearance-and-dynamics reconstruction from visual references. In contrast, our setting treats reconstruction merely as an intermediate step; the agent must subsequently edit the recovered program under an underspecified instruction, ensuring that the final video deviates from the source strictly where the edit and its downstream consequences dictate.

\noindent\textbf{Programmatic Artifacts Editing Benchmarks.}
Code-based editing of visual content has been studied mainly for static artifacts: chart editing~\cite{chartedit} and SVG editing~\cite{svgeditbench,svgenius}. Code editing has also been studied at the function level~\cite{canitedit,codeeditorbench} and the repository level~\cite{swebench}, where coding agents~\cite{sweagent,openhands} achieve strong results given the source code and executable tests. Static visual editing involves no temporal behavior, and general code-editing benchmarks assume access to the source program and measure functional correctness rather than the rendered output. VEDIT-Bench brings these challenges together: the source code is unavailable, and each edit must be propagated through structural, quantitative, and temporal dependencies across the video.

\section{VEDIT Benchmark}
\label{sec:construction}

\begin{figure*}[t]
\centering
\includegraphics[width=\linewidth]{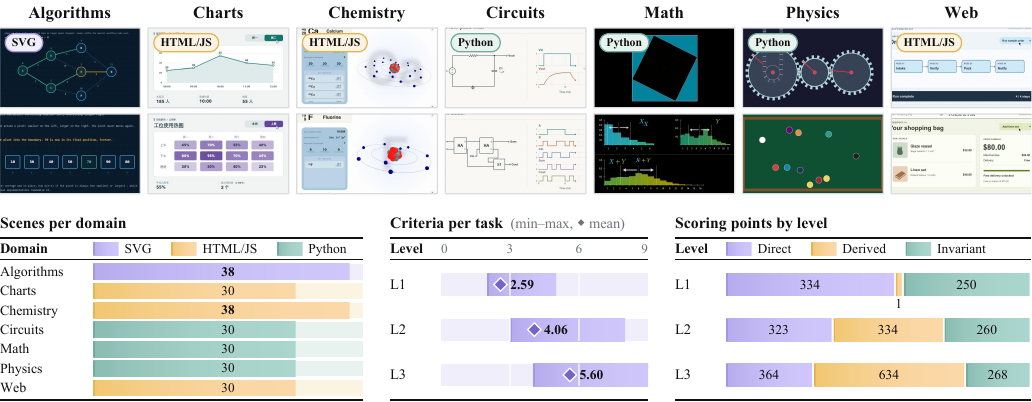}
\caption{\textbf{Dataset examples and statistics.} \emph{Top:} two source scenes from each of the seven domains, with the program family of the domain. \emph{Bottom:} source videos per domain and program family, evaluation criteria per task at each level (min–max and mean), and the number of Direct, Derived, and Invariant scoring points per level over all 678 tasks.}
\label{fig:dataset}
\end{figure*}

\subsection{Task Definition}

In video-conditioned programmatic editing, an agent must modify a programmatic video by writing code, given only its rendered pixels and an under-specified instruction. Unlike standard code editing, the original source code is withheld; unlike pixel-space video editing, the output must be an executable program rather than synthesized frames.

Formally, each benchmark task provides a source video and a concise natural-language instruction that names a target modification while leaving its downstream effects implicit. The agent must output a self-contained program that, when executed, produces an edited video fulfilling three obligations: (i) realizing the explicitly requested change (\emph{Direct}), (ii) carrying out its unstated consequences across time and coupled elements (\emph{Derived}) and (iii) keeping all unrelated content and motion intact (\emph{Invariant}). We evaluate the rendered video against a hidden reference edit using time-anchored semantic metrics and pixel-level measures.

Every source scene is edited at three levels that hold reconstruction difficulty fixed while expanding the scope of causal propagation, as shown in Figure~\ref{fig:task}. \noindent\emph{L1:} Alters a single label, visual attribute, or isolated state. \noindent\emph{L2:} Alters a local element together with directly linked elements, such as a numeric value, its visual encoding, and a summary readout. \noindent\emph{L3:} Alters a governing rule, threshold, or structural relation, requiring later states to be dynamically re-derived.

\subsection{Benchmark Construction}

VEDIT-Bench comprises 678 editing tasks derived from 226 executable source videos across seven domains, as shown in Figure~\ref{fig:dataset}, with each source video instantiated at all three difficulty levels. Source programs span three programmatic paradigms, each paired with a dedicated renderer: SVG animations for algorithms; HTML/JavaScript for charts, chemistry, and web interfaces; and Python for mathematics, physics, and circuits.

\emph{(1) Source Collection.}
We use the frontier LLM, \ie GPT-6-Astra~\cite{gpt6astra}, to mine public repositories pairing generative code with rendered video outputs, filtering for deterministic execution and code quality. It samples 30--38 programs per domain and renders them into reference clips via their respective renderers. A program is admitted into the source pool only if it supports valid edits across all levels.

\emph{(2) Task Synthesis.}
For each source program and difficulty level, the LLM samples a level-specific editing operation, modifies the source code accordingly, renders the ground-truth target video using the corresponding renderer, and generates the editing instruction alongside a set of verification criteria. Criterion density scales with task complexity to verify propagated state changes: the 678 tasks encompass on average 2.59, 4.06, and 5.60 criteria per task for L1, L2, and L3, respectively.

\emph{(3) Audit and Refinement.}
In an isolated session, the LLM audits the instructions and criteria while the MLLM, \ie Opus~5.5~\cite{claudeopus55}, validates the rendered target video against three core constraints:
\emph{(i) Difficulty separation}: edits must strictly adhere to the designated difficulty tier, where L2 modifies a localized group of coupled elements and L3 alters underlying governing rules or relational constraints.
\emph{(ii) Criterion verifiability}: every criterion must be visually decidable solely from the video stream.
\emph{(iii) Instruction specificity with latent intent}: instructions must unambiguously specify target entities and primary actions without explicitly enumerating downstream physical or logical consequences.
Tasks failing any constraint undergo iterative refinement and re-rendering for up to three rounds, after which persistent failures are resampled or discarded. Finally, human annotators verify that every instruction is unambiguous and visually grounded, and that the target video faithfully executes both the primary edit and its dependent state updates.

\subsection{Dataset Statistics}
\label{sec:data-stats}
In VEDIT-Bench, we curate 226 scenes and retain three instructions per scene, yielding 226 source clips and 678 reference edits. These tasks are evenly distributed across difficulty levels (226 per level) and span seven domains—114 each for algorithms and chemistry, and 90 for each of the remaining five. Because all three tasks within a scene share the same source clip, forming a triplet for comparisons.

Figure~\ref{fig:dataset} illustrates the distribution of scoring points across levels: L1 includes 334 Direct, 1 Derived, and 250 Invariant points; L2 comprises 323 Direct, 334 Derived, and 260 Invariant points; and L3 contains 364, 634, and 268 points, respectively.

\begin{figure*}[t]
\centering
\includegraphics[width=\linewidth]{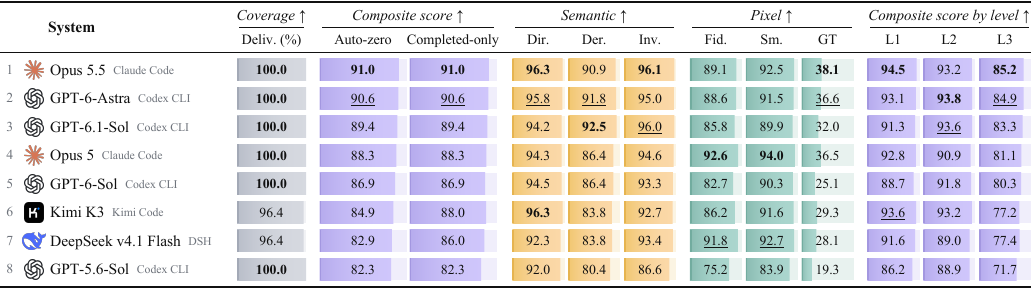}
\caption{
\textbf{Main evaluation results of eight frontier coding agents}. We report delivery coverage (Deliv.), overall composite scores under both Auto-zero and Completed-only protocols, semantic metrics (Dir., Der., Inv.), pixel-level metrics (Fid., Sm., GT), and per-level composite scores (L1–L3 levels). The semantic, pixel-level and per-level composite scores are computed under the Completed-only setting. \textbf{Bold} and \underline{underlined} values denote the best and second-best results per column, respectively.
}
\label{fig:leaderboard}
\end{figure*}

\subsection{Evaluation Metrics}
Conventional global similarity metrics conflate distinct failure modes in programmatic video editing, such as failing to execute the target edit, violating its temporal consequences, or corrupting unrelated content. To address this limitation, we evaluate candidate outputs along three semantic dimensions, \ie, Direct, Derived, and Invariant, and three complementary pixel-level measures, \ie, Fidelity, Smoothness, and GT improvement.
\subsubsection{Semantic-level Evaluation Metrics}
For semantic evaluation metrics, each task is associated with observable, time-anchored scoring points partitioned into \emph{Direct}, \emph{Derived} and \emph{Invariant} metrics. 

\emph{(1) Direct.} The Direct metric verifies whether the explicitly requested object, attribute, or state change is executed at the specified time, \eg, sliding a gear out of mesh. 

\emph{(2) Derived.} The Derived metric verifies whether unstated, rule-governed consequences propagate consistently over time, \eg, halting a disengaged gear's pointer and resuming rotation upon re-engagement. 

\emph{(3) Invariant.} The Invariant metric verifies that no unintended changes occur to unrelated objects, layouts, styles, or motion patterns, \eg, disturbing independent gears.

\subsubsection{Pixel-level Evaluation Metrics}
For pixel-level evaluation, each output is compared against both the source video and the ground-truth reference edit at matched relative timestamps across three complementary dimensions: \emph{Fidelity}, \emph{Smoothness}, and \emph{GT Improvement}. We define the \emph{edit region} as the spatiotemporal area in which the source video and the reference edit differ.

\emph{(1) Fidelity.} The Fidelity metric quantifies how accurately the candidate preserves the local structure and color appearance of the source video outside the edit region, while accounting for any baseline discrepancies already present in the reference edit.

\emph{(2) Smoothness.} The Smoothness metric evaluates temporal consistency by comparing the motion dynamics of the candidate against those of the reference edit, penalizing unnatural freezes, abrupt motion bursts, high-frequency flicker, and deviations in pacing or duration.

\emph{(3) GT Improvement.} The GT Improvement metric measures the proportion of the visual gap between the source video and the reference edit that is closed by the candidate within the edit region, scaled from 0 (identical to the source) to 100 (identical to the reference).

All three metrics are normalized to a 0--100 scale. Notably, because a semantically valid edit rendered in an alternative visual style may achieve a high Direct score despite a low GT Improvement score, GT Improvement is assigned a conservative weight in the composite formulation to avoid unduly penalizing stylistic diversity.

\begin{figure*}[t]
\centering
\includegraphics[width=\linewidth]{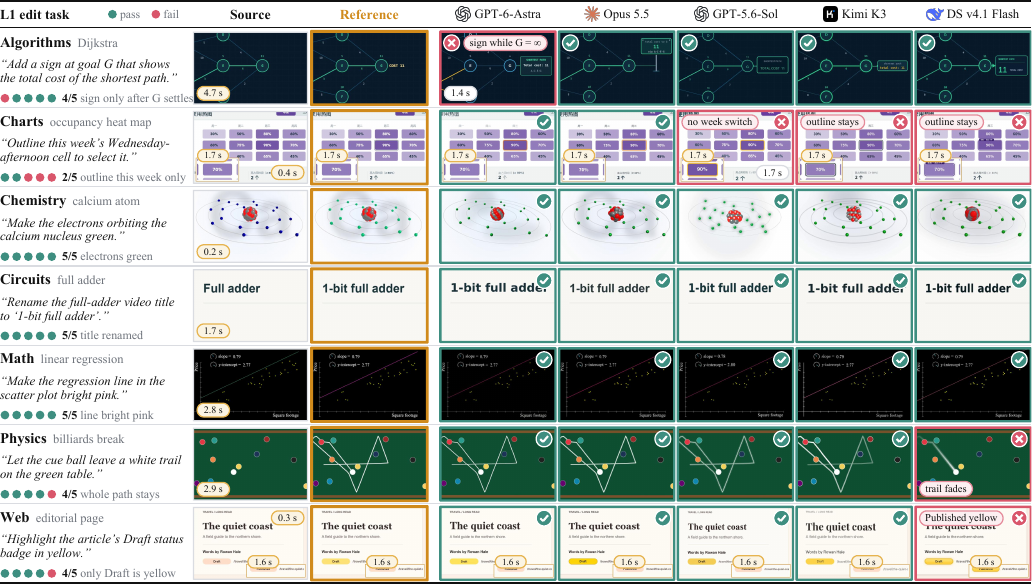}
\caption{\textbf{Qualitative comparison on L1 tasks across seven domains.} We compare rendered edited outputs from five coding agents against the source video and ground-truth reference edit. Dots on the left denote rubric pass/fail counts (\textcolor{teal}{$\bullet$} pass, \textcolor{magenta}{$\bullet$} fail), while pink callouts highlight typical failure modes—such as premature state updates (Dijkstra), persistent selection outlines across weeks (occupancy heat map), fading trails (billiards break), and over-highlighted badges (editorial page).}
\label{fig:l1}
\end{figure*}

\begin{figure*}[t]
\centering
\includegraphics[width=\linewidth]{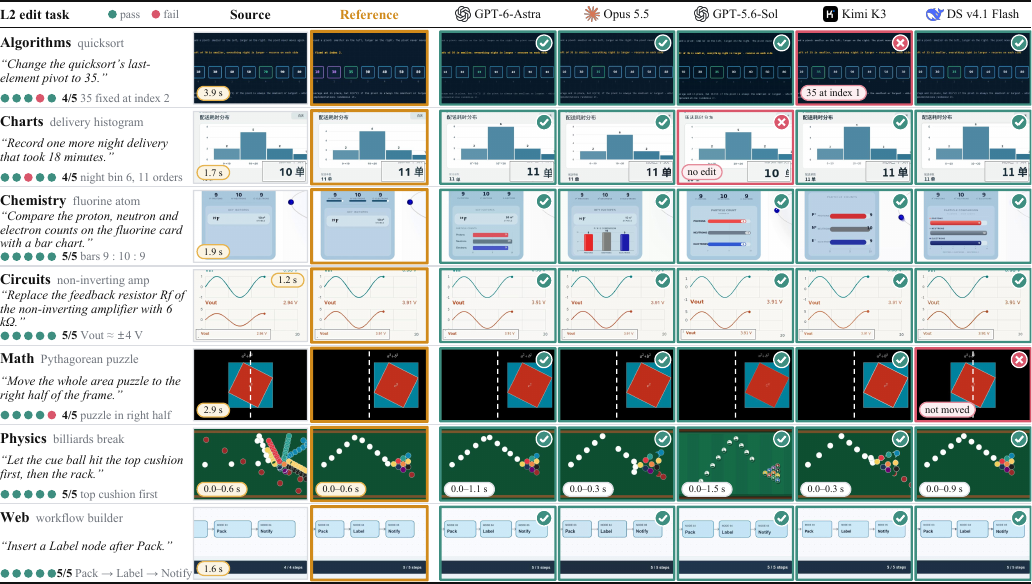}
\caption{\textbf{Qualitative comparison on L2 tasks across seven domains.} Each task requires executing an explicit local edit alongside its immediate quantitative, structural, or physical consequences (e.g., updating histogram bar heights with order totals, or re-routing cue-ball trajectories via the top cushion). While top agents coordinate coupled updates reliably, weaker models omit edits (delivery histogram), misplace pivot indices (quicksort), or fail to translate entire geometric assemblies (Pythagorean puzzle).}
\label{fig:l2}
\end{figure*}

\begin{figure*}[t]
\centering
\includegraphics[width=\linewidth]{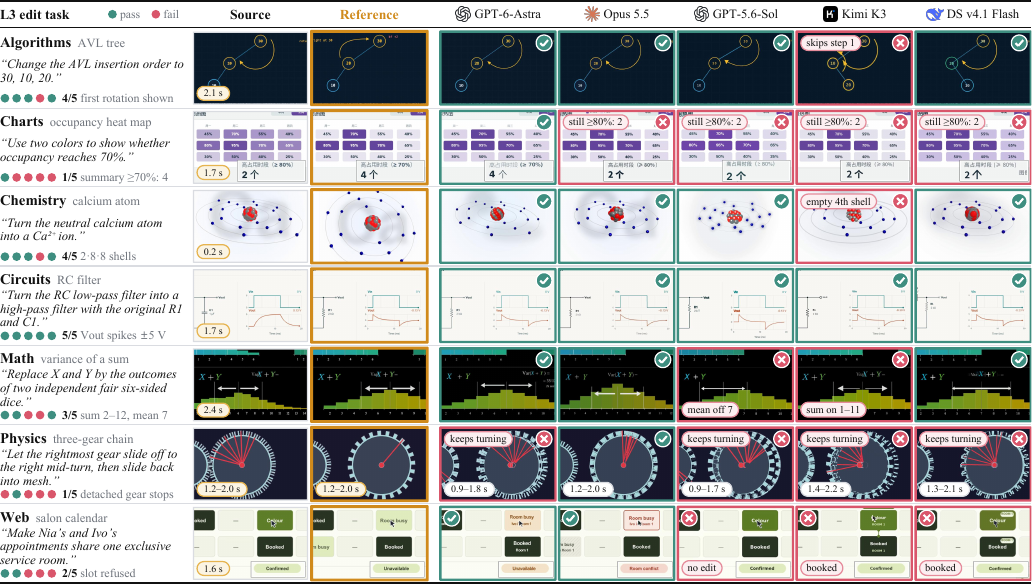}
\caption{\textbf{Qualitative comparison on L3 tasks across seven domains.} Unlike L1 and L2, L3 instructions modify underlying physical, mathematical, or logical rules—such as AVL rotation order, electron shell occupancy ($\text{Ca}^{2+}$), high-pass filter response, or mechanical gear disengagement. Most agents execute surface-level visual edits but fail to propagate downstream causal behavior, e.g., keeping the detached rightmost gear rotating or miscounting occupancy thresholds.}
\label{fig:l3}
\end{figure*}

\subsubsection{Composite Score}

To provide a unified assessment, we combine the six normalized metrics into a single composite score consisting of two components: the \emph{Edit Term} and the \emph{Quality Factor}.

\emph{(1) Edit Term.}
The Edit Term measures instruction-following accuracy and causal reasoning by combining Direct and Derived. When Derived criteria are available, the two metrics contribute 55\% and 45\%, respectively. Otherwise, the Edit Term is determined solely by Direct. This design ensures that preserving the original video cannot compensate for failing to implement the requested modification.

\emph{(2) Quality Factor.}
The Quality Factor is a bounded multiplicative factor ranging from 0.75 to 1. It evaluates the preservation of unrelated content and the overall visual and temporal quality of the output through Invariant, Fidelity, Smoothness, and GT Improvement. Its bounded contribution allows it to penalize unintended changes without overshadowing the execution of the requested edit.

Further details regarding the evaluation metrics and their formal mathematical definitions are in the Appendix.
\section{Experimental Setup}
\label{sec:experiments}

\begin{figure*}[t]
\centering
\includegraphics[width=\linewidth]{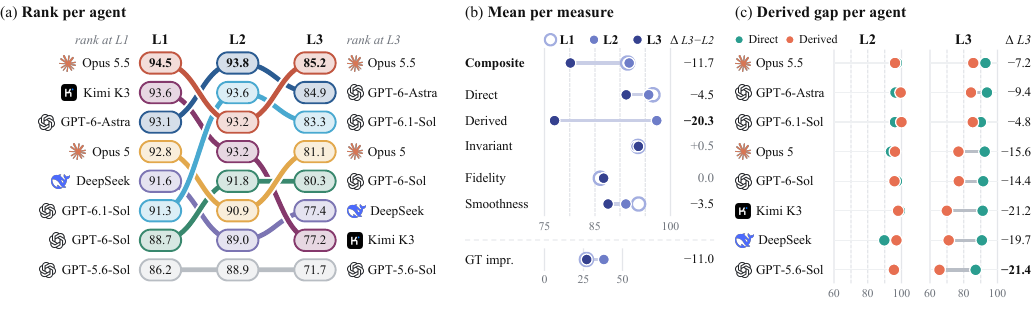}
\caption{\textbf{Performance breakdown across escalating edit levels (completed-only).} \textbf{(a)} Agent rankings and composite scores across L1, L2, and L3, revealing substantial rank shifts as edit complexity increases. \textbf{(b)} Average scores of the evaluation metrics across all eight agents at each level, highlighting the performance degradation from L2 to L3. \textbf{(c)} Per-agent performance comparison between Direct and Derived metrics at L2 and L3.}
\label{fig:levels}
\end{figure*}

\noindent\textbf{Systems.}
We evaluate eight coding agents on the VEDIT-Bench mini-subset, which preserves the full benchmark's coverage across seven domains and pairs each source video with L1, L2, and L3 editing tasks, comprising 84 tasks in total. All agents are configured with high reasoning effort and executed within their native command-line harnesses: GPT-6-Astra~\cite{gpt6astra}, GPT-6.1-Sol~\cite{gpt61sol}, GPT-6-Sol~\cite{gpt6sol}, and GPT-5.6-Sol~\cite{gpt56sol} via Codex CLI; Opus~5.5~\cite{claudeopus55} and Opus~5~\cite{claudeopus5} via Claude Code; Kimi~K3~\cite{kimik3} via Kimi Code; and DeepSeek v4.1 Flash~\cite{deepseekv41flash} via DeepSeekHarness.

\noindent\textbf{Protocol.}
For each editing task, the coding agent receives the source video along with an underspecified editing instruction. To complete the task, the agent must infer the implicit editing intent and synthesize a standalone program that reconstructs the scene from scratch and executes the target edit, without referencing the source video or reusing its original frames. Throughout execution, the ground-truth source code, reference edited video, and evaluation rubrics remain withheld, and the final rendered video is assessed across multiple evaluation criteria.

\noindent\textbf{Judging.}
For each semantic-level criterion, we employ GPT-6.1-Terra~\cite{gpt61terra} at high reasoning effort within the Codex CLI as our VLM judge. Equipped with local video-inspection tools, the judge first reviews full-duration overviews of the source, reference edited, and rendered edited videos, and then examines consecutive frames from the reference and rendered videos around task-relevant transitions. It adaptively selects timestamps, temporal intervals, and enlarged views or crops as needed to compare corresponding semantic phases while tolerating minor timing variations. By contrast, pixel-level metrics are computed programmatically from the rendered edited videos.

\noindent\textbf{Aggregation.}
Scores are calculated per video and averaged with equal weight across tasks. \textit{Auto-zero} accounts for every registered task, assigning a zero whenever an output is missing. \textit{Completed-only} averages exclusively over delivered videos; consequently, for the composite metric, Auto-zero equals Completed-only multiplied by delivery rate.
\section{Results and Analysis}
\label{sec:results}

\begin{figure*}[t]
\centering
\includegraphics[width=\linewidth]{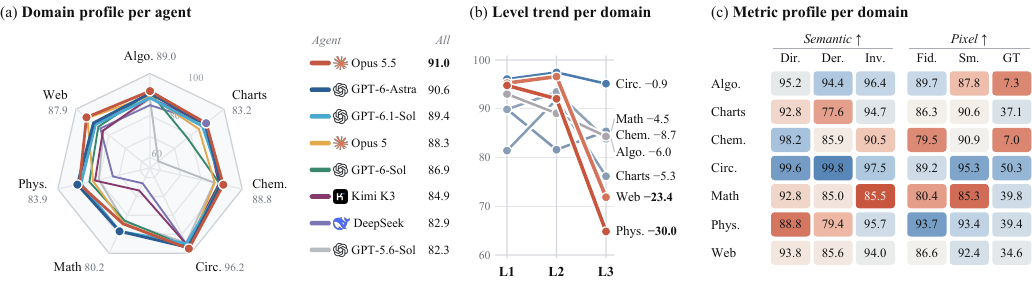}
\caption{\textbf{Domain performance and metric profiles.}
\textbf{(a)} Auto-zero composites by agent and domain; axis labels give the mean over agents.
\textbf{(b)} Completed-only composites averaged over agents at each level, labeled with the L3$-$L1 change.
\textbf{(c)} Completed-only means of semantic and pixel-level metrics per domain.}
\label{fig:domains}
\end{figure*}

\begin{figure*}[t]
\centering
\includegraphics[width=\linewidth]{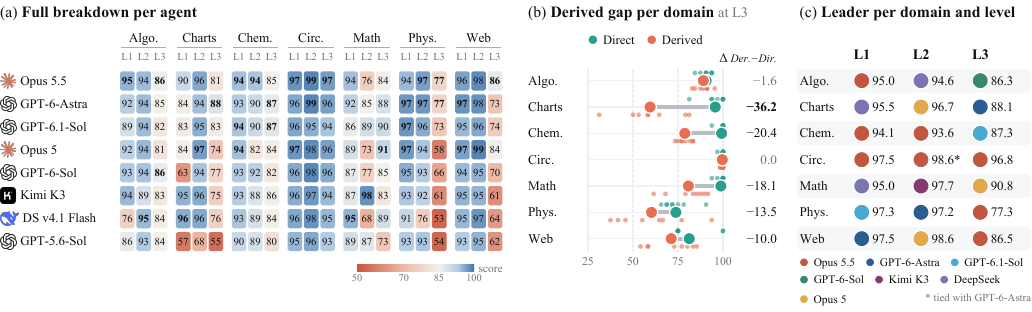}
\caption{\textbf{Performance by domain and editing level.}
\textbf{(a)} Completed-only composites by agent, domain, and level; bold marks column maxima.
\textbf{(b)} Direct and Derived scores at L3 by domain.
\textbf{(c)} Best agent and its composite for each domain and level.}
\label{fig:breakdown}
\end{figure*}

\noindent\textbf{Overall Results.}
As Figure~\ref{fig:leaderboard} shown, Opus~5.5 achieves the highest composite score, followed by GPT-6-Astra and GPT-6.1-Sol. Direct scores are uniformly high across agents, whereas Derived scores spread more widely. Agents excel at local modifications and immediate coupled updates, but performance drops when instructions alter underlying scene rules or structural relationships. As difficulty increases, inter-agent gaps widen and relative rankings shift. The primary bottleneck is propagating edits across dependent states and events: models often apply the explicit visual change while leaving downstream behaviors outdated.

\noindent\textbf{Analysis on Different Editing Levels.}
Figures~\ref{fig:l1}--\ref{fig:l3} show qualitative examples at each level, and Figure~\ref{fig:levels} quantifies how agent strengths shift across L1, L2, and L3. For L1 tasks shown in Figure~\ref{fig:l1}, top systems cluster tightly with high scores; Opus~5.5~\cite{claudeopus55} leads alongside Kimi~K3~\cite{kimik3}, while DeepSeek v4.1 Flash~\cite{deepseekv41flash} and GPT-6.1-Sol place mid-pack, showing limited differentiation on local edits. For L2 tasks shown in Figure~\ref{fig:l2}, overall performance holds steady with comparable Direct and Derived scores, indicating reliable coordination of immediate dependencies. However, as shown in Figure~\ref{fig:l3}, L3 triggers a sharp decline across all systems. Rankings shift as GPT-6-Astra and GPT-6.1-Sol rise to second and third behind Opus~5.5, while Kimi~K3 and DeepSeek v4.1 Flash~\cite{deepseekv41flash} drop toward the bottom, demonstrating that local-editing strength does not generalize to broader structural transformations. Across all agents, including L3 leaders, Derived correctness drops much more steeply than Direct correctness, while Invariant preservation and fidelity remain stable. Ultimately, while current agents handle local and immediate updates well, propagating global rule changes remains a capability gap.

\noindent\textbf{Analysis on Different Domains.}
Figures~\ref{fig:domains} and~\ref{fig:breakdown} break performance down by domain. Difficulty varies markedly across domains: circuits are nearly saturated, whereas mathematics, charts, and physics are the hardest and separate agents the most. Part of the mathematics gap reflects missing deliveries, as Kimi~K3 and DeepSeek v4.1 Flash each miss one mathematics scene. The level effect is also domain-dependent: from L1 to L3, physics and web lose 30.0 and 23.4 points, whereas circuits lose only 0.9. Charts and physics also have the lowest Derived scores overall. At L3, the decline is concentrated in Derived criteria. On charts, chemistry, and mathematics, mean Direct stays above 95, yet Derived trails it by 36.2, 20.4, and 18.1 points, respectively, while in physics even Direct drops to 73.8. No single agent dominates: across the 21 domain--level cells, Opus~5.5 leads eight and six other agents share the rest, although every L3 cell is led by one of the five strongest agents overall.

\noindent\textbf{Analysis on Bottlenecks of Current Agents.}
The L3 results expose a recurring limitation in translating editing intent into code: agents often implement the explicitly named visual change but fail to encode its causal consequences on the rest of the scene.
Qualitative examples illustrate this failure mode, as shown in Figure~\ref{fig:l3}.
In the three-gear task, most agents keep the detached rightmost gear turning rather than stopping it while it is out of mesh; only Opus~5.5 stops it.
Similarly, in the occupancy heat-map task, most agents keep the original 80\% threshold instead of the requested 70\%, so the summary still counts 2 cells instead of 4.
In both cases, the edited scene still follows the original rule rather than the one implied by the instruction.
Together, these trends demonstrate that incomplete dependency inference and state-transition revision remain bottlenecks.

\noindent\textbf{Discussion.}
The empirical results validate our benchmark's progression from local modifications (L1) and coupled updates (L2) to rule-dependent transformations (L3), which require inferring broader consequences and recomputing scene behavior.
The uniform performance decline and substantial ranking shifts at L3 confirm its increased difficulty and discriminative value, while the concentration of errors in dependent updates explains this gap.
Ultimately, the benchmark evaluates whether an agent can translate editing intent into a coherent program that reflects modified rules, offering an effective metric for reliable code-based video editing and highlighting key areas for improvement.
\section{Conclusion}
We introduce VEDIT-Bench, a unified benchmark for evaluating coding agents on video-conditioned programmatic editing under undescribed instructions, spanning 678 tasks and 226 scenes across three escalating difficulty levels (L1–L3). By pairing Direct, Derived, and Invariant semantic metrics with pixel-level measures, our benchmark decouples explicit edit execution from implicit causal propagation and background preservation. Extensive evaluations of eight frontier coding agents reveal that current models still struggle to infer latent modification intents from underspecified instructions and translate them into executable video-editing code.

\section*{LLM Usage Statement}
We used ChatGPT (OpenAI) solely for language polishing,
grammar correction, and improving the clarity of the
manuscript. All scientific ideas, methodologies,
experiments, analyses, and conclusions were developed
and verified by the authors.

{
    \small
    \bibliographystyle{ieeenat_fullname}
    \bibliography{main}
}

\ifwithsupp
\clearpage
\maketitlesupplementary
\appendix

\noindent
This supplementary material is organized into three sections. \cref{sec:metrics} details the evaluation metrics and the composite score, \cref{sec:implementation} describes the experimental implementation, and \cref{sec:gallery} presents representative video results across all seven domains.

\section{Evaluation Metrics}
\label{sec:metrics}

We evaluate delivered videos using six measures on a 0--100 scale where higher is better, combining the applicable measures into one composite score. Three measures are semantic (Direct, Derived, Invariant) and three are computed from pixels (Fidelity, GT~Improvement, Smoothness). We write $\src$ for the source video, $\gt$ for the reference edit and $\cand$ for the video produced by an agent.

\subsection{Semantic-level Measures}
\label{sec:semantic}

To enable fine-grained and interpretable evaluation, each editing task is annotated with a curated set of atomic, visually verifiable criteria that can be directly assessed by inspecting the generated video. Rather than relying on a single holistic score, these criteria are partitioned into three categories: \emph{Direct} ($\Dir$), which evaluates the explicit edits directly specified by the instruction; \emph{Derived} ($\Der$), which assesses the implicit physical, causal, or contextual consequences triggered by the edit (e.g., object interactions); and \emph{Invariant} ($\Inv$), which verifies the preservation of unaffected content and dynamics. Each category contains task-specific criteria tailored to the scene, with Derived criteria omitted when inapplicable. A Vision-Language Model (VLM) serves as an automated judge, inspecting the source, reference, and rendered videos, including consecutive frames around relevant transitions. It assigns each criterion an integer score from 0 to 4, reflecting the degree of visual compliance:
\begin{itemize}
\item 0: not done, or the opposite of what is required;
\item 1: only a faint sign of the required change;
\item 2: partly correct, but something important is wrong;
\item 3: mostly correct, with a small visible defect;
\item 4: fully correct.
\end{itemize}
For a given task, the VLM judge rates each sub-criterion on a $0\text{--}4$ integer scale. We first compute the task-level score for each group by averaging the ratings of its constituent sub-criteria and multiplying by $25$ to map the result onto a $0\text{--}100$ scale:
\begin{equation}
\Dir,\ \Der,\ \Inv \;=\; 25\times(\text{avg. rating of the group}).
\label{eq:semantic}
\end{equation}
Once the per-task scores are obtained, we average them across all tasks within a domain to report domain-level performance. If a task has no \emph{Derived} criteria, $\Der$ is marked as not applicable and omitted from both the task-level composite calculation and the domain-level average.

To make this hierarchical scoring concrete, consider a representative Level-3 task from the \emph{Algorithms} domain (\eg, Dijkstra relaxation): the editing instruction is \emph{``Remove all edges connected to C and recompute the shortest path from A to G.''} Although the prompt is concise, evaluating it requires verifying both the immediate structural modification and its downstream algorithmic consequences:
\begin{itemize}
    \item \textbf{Direct ($\Dir$):} Captures the sole explicit structural edit requested---whether node C remains in the graph with all incident edges removed ({Dir.~1}). The VLM's $0\text{--}4$ rating on this single item directly determines $\Dir$ ($25 \times \text{rating}$).
    \item \textbf{Derived ($\Der$):} Decomposes the implicit algorithmic cascade into four sequential sub-criteria across the video timeline: (1)~whether C's distance readout remains $\infty$ throughout every execution stage ({Der.~1}); (2)~whether node D's distance entry updates from $\infty$ to $9$ immediately after node B is processed ({Der.~2}); (3)~whether the final highlighted path highlights edges A--B, B--D, and D--G in sequence ({Der.~3}); and (4)~whether the final distance entry for G settles at $13$ ({Der.~4}). The VLM scores each of the four \emph{Derived} sub-criteria on the $0\text{--}4$ scale; their arithmetic mean is then multiplied by $25$ to yield the task's overall $\Der$ score.
    \item \textbf{Invariant ($\Inv$):} Uses one criterion to verify that source A, destination G, and unaffected edge weights are preserved, while node highlighting and the distance table continue to update with each step ({Inv.~1}). This criterion applies only to content unaffected by the instruction and its reasonable consequences.
\end{itemize}

For this task, the VLM assigns a $0\text{--}4$ integer score to each sub-criterion ($1$ item in $\Dir$, $4$ in $\Der$, and $1$ in $\Inv$). Within each group, the ratings are averaged and multiplied by $25$ to produce the task-level $\Dir$, $\Der$, and $\Inv$ scores, ensuring proportional credit assignment across explicit edits, causal reasoning, and scene preservation.

\subsection{Pixel-level Measures}
\label{sec:pixel}

\paragraph{Preparation.}
For Fidelity and GT~Improvement, all three videos are sampled at 33 uniformly spaced relative timestamps, selecting the nearest frames from the first to the last frame, and resized to $256\times144$ pixels. This aligns videos by normalized time rather than explicitly matching semantic stages. We define the \emph{edit region} as the set of pixels exhibiting noticeable discrepancies between the source and reference videos---specifically, where any color channel differs by more than $16/255$---dilated by a 1-pixel margin (\cref{fig:mask}). Ideally, the candidate output should match the reference inside the edit region while preserving source content outside it.

To evaluate frame-level correspondence between two frames $X$ and $Y$, we formulate a per-pixel similarity metric $\kappa(X,Y) \in [0,1]$ that jointly captures structural and chromatic fidelity:
\begin{equation}
\kappa(X,Y)=0.75\cdot\operatorname{SSIM}(X,Y)+0.25\cdot e^{-5\,\lvert X-Y\rvert},
\label{eq:kappa}
\end{equation}
where $\operatorname{SSIM}$ denotes local grayscale structural similarity clamped to $[0,1]$, and $\lvert X-Y\rvert$ is the mean absolute color difference over the three channels, normalized to $[0,1]$.

\begin{figure}[t]
\centering
\includegraphics[width=\linewidth]{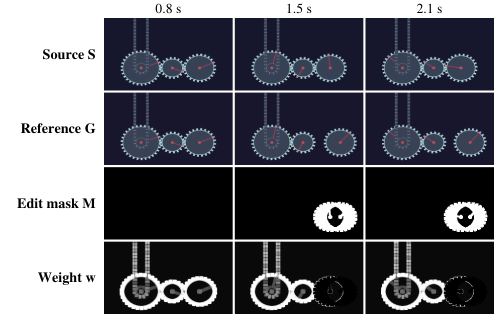}
\caption{Visualization of the edit region across three timestamps. White pixels denote the edit region where the source and reference diverge, evaluated via GT~Improvement. The remaining regions are evaluated via Fidelity with spatial weights $w$ (higher intensity indicates larger weight), prioritizing lines, text, and edges over uniform backgrounds.}
\label{fig:mask}
\end{figure}

\paragraph{Fidelity.}
Fidelity evaluates background preservation strictly \emph{outside} the edit region. At each pixel, the candidate incurs a penalty whenever its similarity to the source drops below that of the reference:
\begin{equation}
\mathcal{L}_{\text{fid}}=\max\bigl(0,\;\kappa(\src,\gt)-\kappa(\src,\cand)\bigr).
\end{equation}
The overall Fidelity score is computed as the weighted average of $1-\mathcal{L}_{\text{fid}}$ across all unedited pixels and all 33 sampled frames:
\begin{equation}
\Fid = 100\times\frac{\sum w\,(1-\mathcal{L}_{\text{fid}})}{\sum w},
\label{eq:fidelity}
\end{equation}
where the spatial weight $w$ assigns higher importance to high-frequency structures (e.g., lines, text, and edges) in the source video than to flat backgrounds. Furthermore, if the candidate's aspect ratio deviates from that of the reference, $\Fid$ is penalized by multiplying it by their aspect ratio quotient (smaller over larger).

\paragraph{GT Improvement.}
GT~Improvement measures edit realization strictly \emph{inside} the edit region, quantifying the degree to which the candidate transitions from the source toward the reference. It averages two complementary normalized metrics. The first measures perceptual similarity gain: let $H$ and $B$ denote the mean percentage similarity $\kappa$ to the reference for the candidate and the unedited source, respectively. The second measures chromatic error reduction: let $e_{\cand}$ and $e_{\src}$ denote the cumulative $\ell_1$ color error relative to the reference for the candidate and the source. Formally,
\begin{equation}
\GTI = \frac12\cdot\underbrace{100\cdot\frac{H-B}{100-B}}_{\text{similarity gain}}
+\frac12\cdot\underbrace{100\cdot\Bigl(1-\frac{e_{\cand}}{e_{\src}}\Bigr)}_{\text{error reduction}},
\label{eq:gt}
\end{equation}
with both terms clamped to $[0, 100]$. If $B$ is numerically indistinguishable from $100$, only the error-reduction term is used. This metric requires a nonempty reference edit region. By construction, an unaltered copy of the source yields $\GTI=0$, whereas an exact match to the reference achieves $\GTI=100$. Note that a valid edit rendered with a distinct visual style from the reference may still obtain a low $\GTI$; consequently, this metric is assigned a relatively low weight in our composite score.

\paragraph{Smoothness.}
Smoothness uses all decoded frames, resized to $256\times144$ pixels. We quantify inter-frame visual changes and aggregate these motion dynamics across 12 uniform temporal stages for both the candidate and reference videos. Starting from a maximum score, the candidate incurs penalties for five types of temporal deviations relative to the reference, each normalized to $[0,1]$:
\begin{itemize}
\item $a$: stage-wise discrepancy in motion magnitude;
\item $z$: excess freezing during stages where the reference exhibits motion;
\item $b$: excess variation in motion magnitude across temporal stages;
\item $c$: reduced effective frame rate or excess timestamp irregularity;
\item $\ell$: excess temporal flickering relative to the reference, measured by transient pixel-color reversals across consecutive frames.
\end{itemize}
Three multiplicative scaling factors further modulate the score: $r$ is the candidate-to-reference ratio of non-frozen time fractions within reference-active stages, capped at $1$; $d$ is the duration ratio (shorter over longer); and $h$ measures excess global color flashes relative to the reference, allowing a small tolerance. The overall Smoothness score is formulated as:
\begin{equation}
\begin{aligned}
\Smo = 100&\times(1-0.35a-0.20z-0.10b-0.15c-0.20\ell)\\
&\times(0.45+0.55\,r)\times d^{1/4}\times e^{-2.5h}.
\end{aligned}
\label{eq:smooth}
\end{equation}
The score is clamped to $[0,100]$. An exact copy of the reference, including its timing, achieves $\Smo=100$.

\subsection{Composite Score}
\label{sec:composite}

The composite score integrates an \emph{edit term} $E$, reflecting task completion accuracy, with a multiplicative \emph{quality factor} $Q$, capturing perceptual and temporal fidelity.

\paragraph{Edit term.} $E$ combines the Direct and Derived criteria with weights of $55\%$ and $45\%$, respectively:
\begin{equation}
E = 0.55\,\Dir+0.45\,\Der ,
\label{eq:edit}
\end{equation}
and simplifies to $E=\Dir$ for tasks lacking Derived score.

\paragraph{Quality factor.} $Q$ aggregates the remaining four auxiliary metrics and is bounded within $[0.75, 1.0]$:
\begin{equation}
Q = 0.75+\frac{0.10\,\Inv+0.05\,\Fid+0.05\,\Smo+0.05\,\GTI}{100}.
\label{eq:quality}
\end{equation}

\paragraph{Overall Score.}
The overall score is defined as the product of the edit term $E$ and the quality factor $Q$:
\begin{equation}
\text{Score}=E\times Q .
\label{eq:score}
\end{equation}
As an exception, returning the source video unchanged yields a total score of $0$, identified by identical decoded frames, dimensions, and relative timestamps.

\paragraph{Averaging over tasks.}
Scores are computed per video and subsequently averaged across tasks with equal weights, excluding inapplicable Derived scores. We report two aggregation protocols: \emph{Auto-zero}, which includes all registered tasks and assigns $0$ to applicable metrics for missing deliveries; and \emph{Completed-only}, which includes only delivered videos.

\section{Implementation Details}
\label{sec:implementation}

\paragraph{Computing infrastructure and model access.}
Our local setup comprises two NVIDIA A100 GPUs dedicated to on-premise workloads and environment execution. Inference for hosted models is offloaded to external provider infrastructure and accessed via official APIs within each agent's native harness through Harbor. Following the main paper, all agents are evaluated under the high reasoning effort.

\paragraph{Information available to the agent.}
During editing, the agent receives only a concise natural-language instruction, with no access to the reference edit or task-specific evaluation criteria; the reference video is strictly reserved for evaluation. The generated program must execute standalone without relying on the source video, nor may it embed or replay original frames. Accordingly, evaluation assesses both the rendered consequences of the requested edit and the preservation of unedited content.

\paragraph{Rendering and evaluation inputs.}
As detailed in the main paper, the benchmark spans SVG animations, HTML/JavaScript scenes, and Python programs, with the final rendered video serving as the unified evaluation input across all systems. For pixel-level metrics, source, reference, and candidate clips are temporally aligned by relative timestamps using the sampling and resizing protocol in \cref{sec:pixel}. Qualitative visual grids follow the same relative-time alignment to enable stage-wise comparison across clips of varying durations.

\section{Qualitative Comparison}
\label{sec:gallery}

The following pages present representative qualitative results spanning all seven domains of our benchmark. To systematically illustrate how model performance scales with task complexity, we showcase one representative scene per domain, evaluated progressively across all three difficulty levels (L1, L2, and L3).

Each figure is dedicated to a single editing task and is organized as a unified spatiotemporal grid for direct side-by-side comparison. The top row displays frames from the unedited source video, while the second row (highlighted by an orange border) presents the ground-truth reference edit. Each subsequent row depicts the generated output of a competing baseline under identical inference settings. To facilitate a fair assessment of temporal consistency and motion alignment, all rows are sampled at identical relative timestamps, with column headers denoting the exact timestamps within the reference clip. Furthermore, the rightmost two columns provide magnified crops of fixed spatial regions across all rows---corresponding to the dashed bounding boxes annotated in the source and reference rows---allowing readers to closely inspect fine-grained visual details, boundary artifacts, and unintended background alterations. Finally, each figure caption explicitly states the text editing instruction alongside the task's Direct, Derived and Invariant criteria, offering a structured guide on which visual cues and physical interactions to scrutinize.

\clearpage
\begin{figure*}[p]
\centering
\pdfbookmark[2]{Algorithms / Four queens (backtracking) / L1}{bm-algorithms__backtracking-n-queens__L1}
\includegraphics[width=\linewidth]{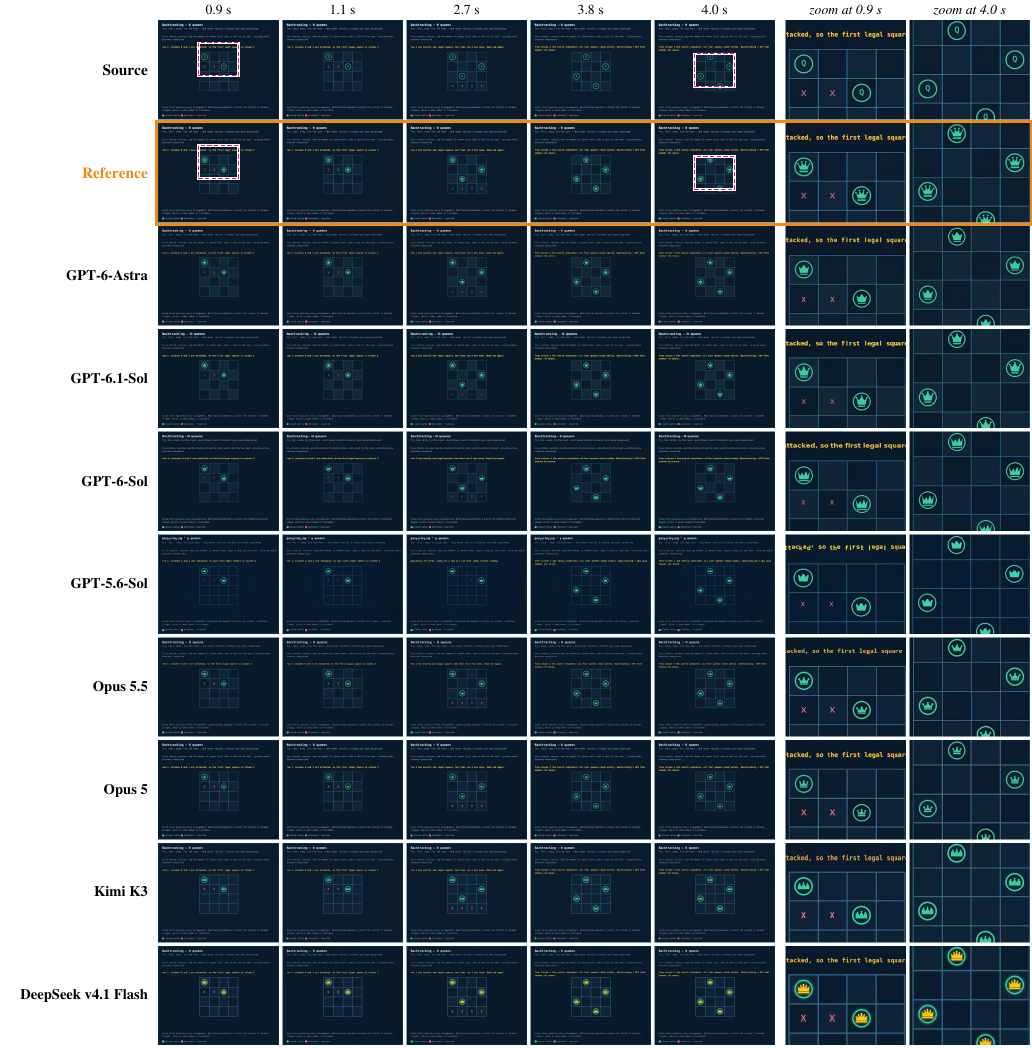}
\caption{\gallerycaptionfont \textbf{Algorithms / Four queens (backtracking) / L1.} \emph{Instruction:} ``Replace the letter Q on the four-queens board with a crown icon.'' \textbf{Direct:} (D1)~During the trial-placement phase, are the positions where queens have been placed on the board shown with green crown-shaped piece glyphs? \textbf{Invariant:} (I1)~At the end of the video, does each of the four rows hold one crown-shaped glyph, located in columns 2, 4, 1 and 3 from the left respectively? (I2)~The four-row, four-column board and the step-by-step display of trial placement and backtracking are kept, and the solution has one queen in each row and each column with no mutual attacks.}
\label{fig:g-algorithms-backtracking-n-queens-L1}
\end{figure*}

\begin{figure*}[p]
\centering
\pdfbookmark[2]{Algorithms / Four queens (backtracking) / L2}{bm-algorithms__backtracking-n-queens__L2}
\includegraphics[width=\linewidth]{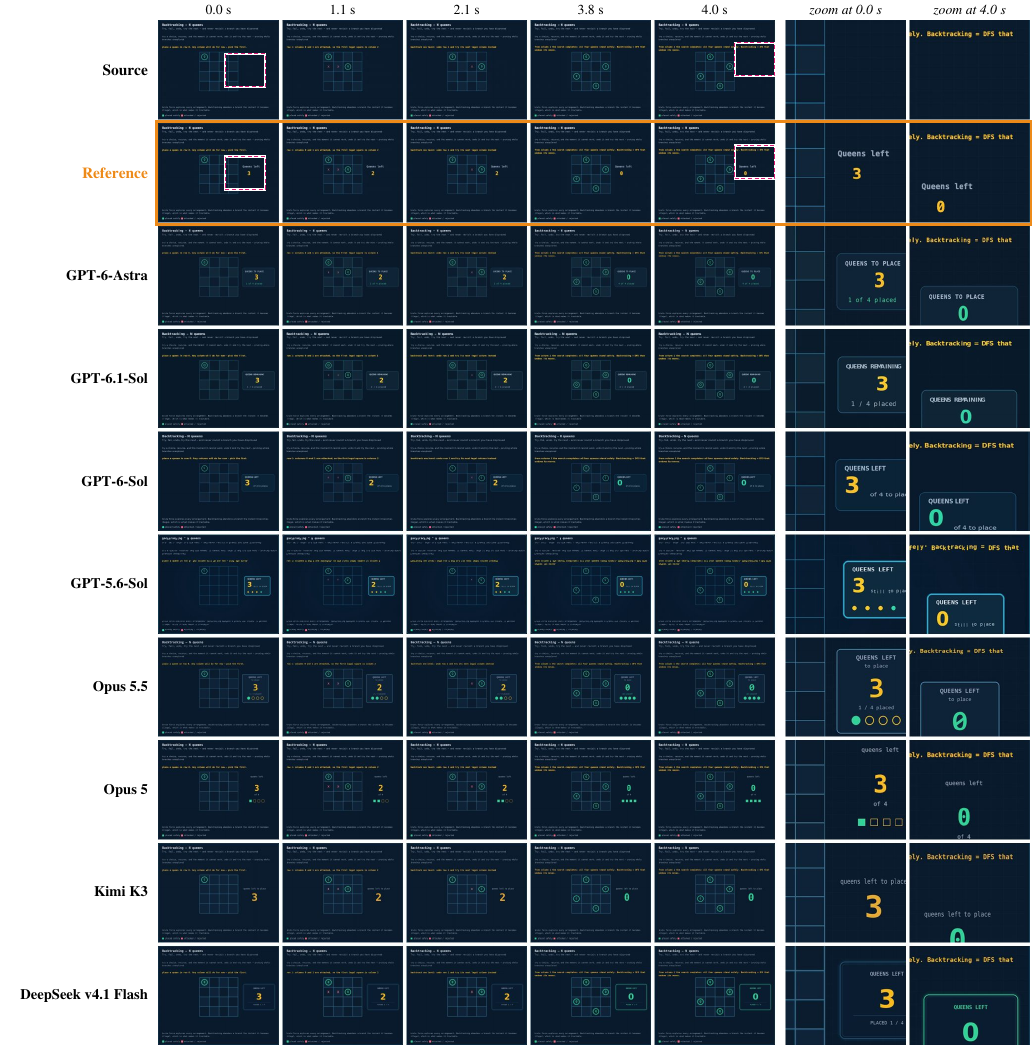}
\caption{\gallerycaptionfont \textbf{Algorithms / Four queens (backtracking) / L2.} \emph{Instruction:} ``Show the number of queens still to be placed beside the four-queens board.'' \textbf{Direct:} (D1)~At the opening, when the board has only one Q in the top-left corner, is the remaining count beside the board 3? \textbf{Derived:} (R1)~When backtracking from three Qs to only one Q in the first row, does the remaining count rise from 1 back to 3? (R2)~When the solution with one Q in each of the four rows appears, does the remaining count become 0? \textbf{Invariant:} (I1)~The four-row, four-column board and the step-by-step display of trial placement and backtracking are kept, and the solution has one queen in each row and each column with no mutual attacks.}
\label{fig:g-algorithms-backtracking-n-queens-L2}
\end{figure*}

\begin{figure*}[p]
\centering
\pdfbookmark[2]{Algorithms / Four queens (backtracking) / L3}{bm-algorithms__backtracking-n-queens__L3}
\includegraphics[width=\linewidth]{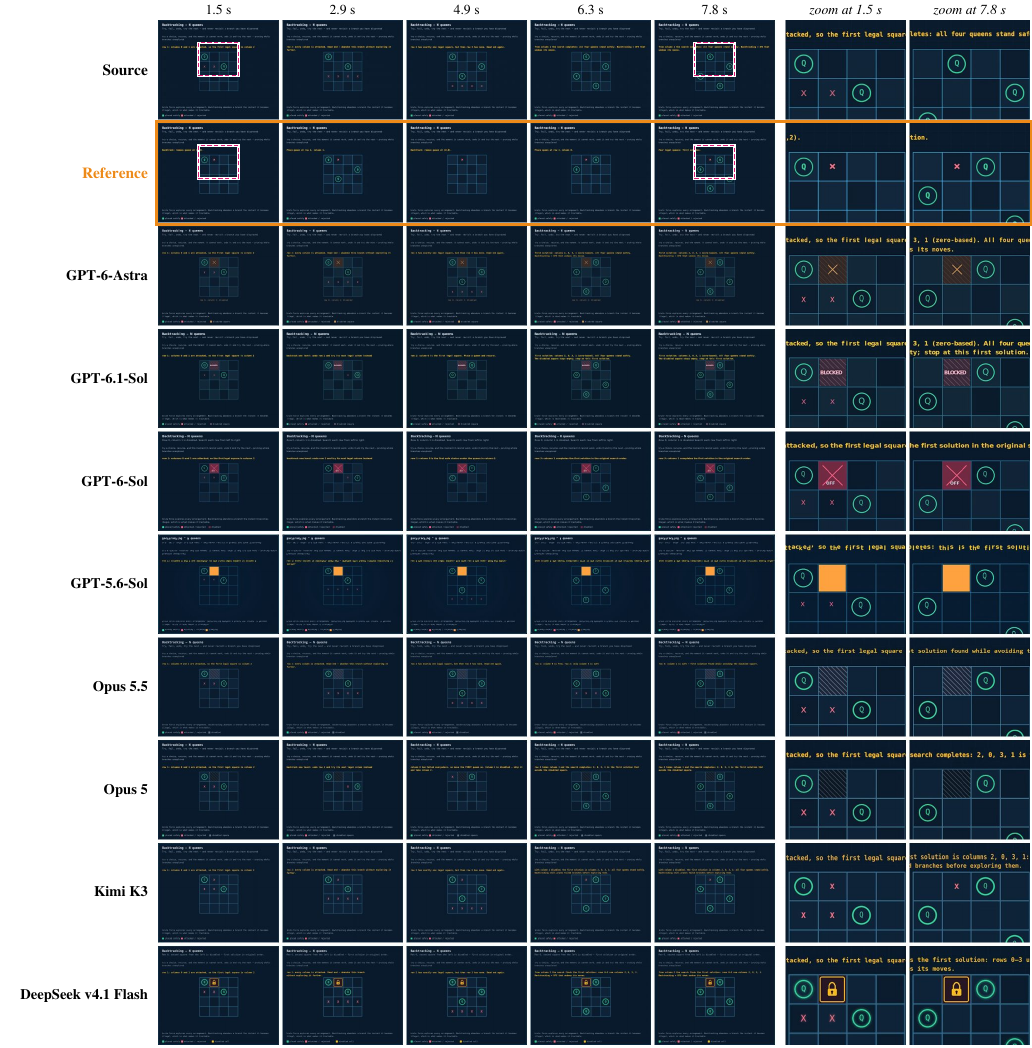}
\caption{\gallerycaptionfont \textbf{Algorithms / Four queens (backtracking) / L3.} \emph{Instruction:} ``Disable the second square from the left in the first row of the four-queens board, and search for the first solution in the original order.'' \textbf{Direct:} (D1)~Does the second square from the left in the first row of the board carry a cross mark from the very opening? (D2)~From trial placement to the solution, does the second square from the left in the first row keep its cross mark throughout, with no Q appearing in that square? \textbf{Derived:} (R1)~After a Q appears in the third square from the left in the second row, does that square become empty again, while the Q in the top-left corner is kept both before and after the backtrack? (R2)~In the solution, do the four Qs from top to bottom fall in squares 3, 1, 4 and 2 from the left respectively? \textbf{Invariant:} (I1)~Is the first Q placed in the top-left corner of the board? (I2)~The four-row, four-column board and the step-by-step display of trial placement and backtracking are kept, and the solution has one queen in each row and each column with no mutual attacks.}
\label{fig:g-algorithms-backtracking-n-queens-L3}
\end{figure*}

\begin{figure*}[p]
\centering
\pdfbookmark[2]{Algorithms / Dijkstra relaxation / L3}{bm-algorithms__dijkstra-relax__L3}
\includegraphics[width=\linewidth]{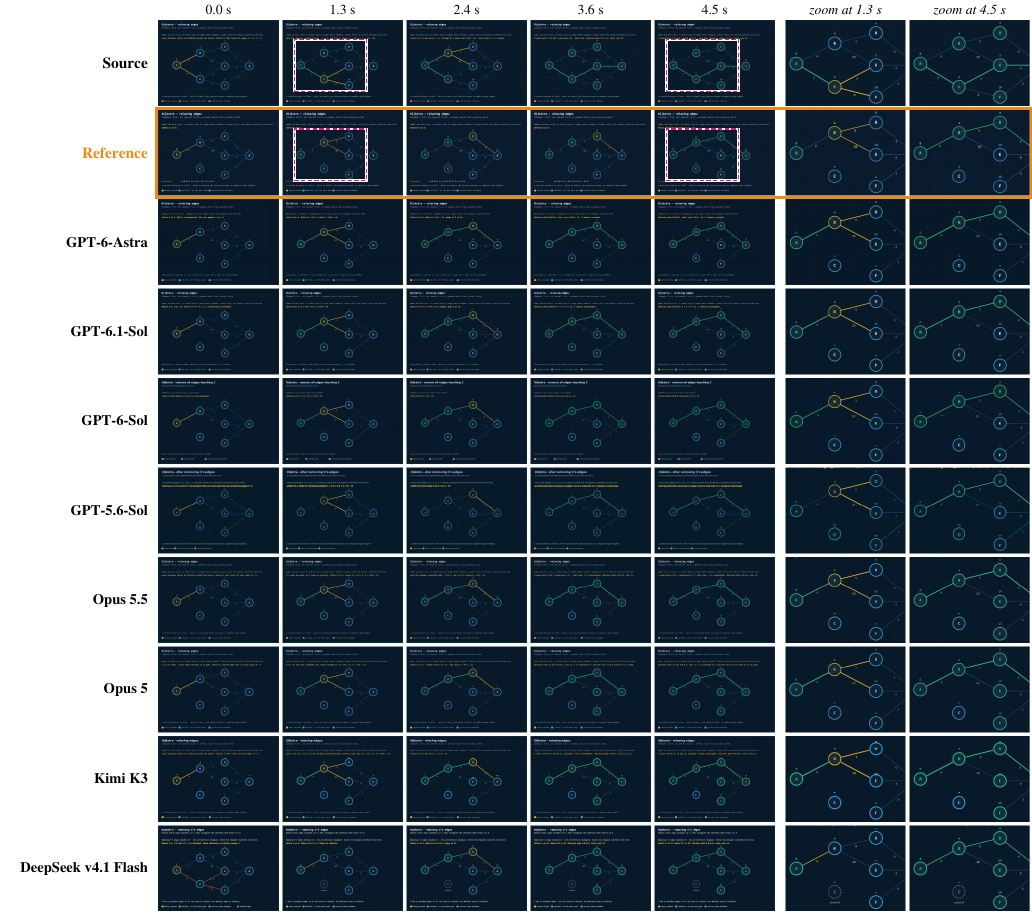}
\caption{\gallerycaptionfont \textbf{Algorithms / Dijkstra relaxation / L3.} \emph{Instruction:} ``Remove all edges connected to C and recompute the shortest path from A to G.'' \textbf{Direct:} (D1)~Is node C still in the graph, but with no edge connected to it? \textbf{Derived:} (R1)~In the Distances row at every stage, is the readout of C always $\infty$? (R2)~When node B has been processed, is the value of D in the Distances row below the graph updated from $\infty$ to 9? (R3)~At the end of the video, are the highlighted edges from A to G, in order, A--B, B--D and D--G? (R4)~At the end of the video, does the G entry of the Distances row below the graph show 13? \textbf{Invariant:} (I1)~Source A, destination G and the edge weights not involved are kept, and the highlighting of the node being processed and the distance table still update with each step.}
\label{fig:g-algorithms-dijkstra-relax-L3}
\end{figure*}

\clearpage

\begin{figure*}[p]
\centering
\pdfbookmark[2]{Charts / Passenger-flow line chart / L1}{bm-charts__01-trend__L1}
\includegraphics[width=\linewidth]{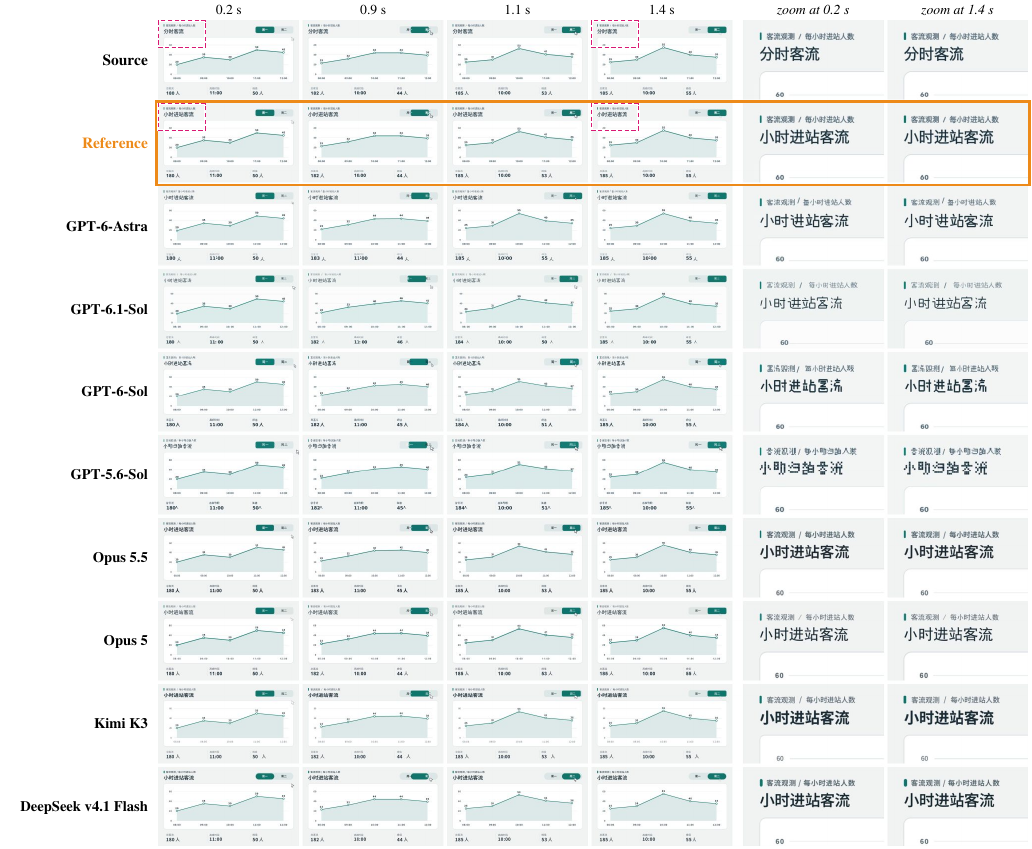}
\caption{\gallerycaptionfont \textbf{Charts / Passenger-flow line chart / L1.} \emph{Instruction:} ``Change the title of the passenger-flow chart to ``Hourly station entries''.'' \textbf{Direct:} (D1)~Does the main title at the upper left of both the Monday and Tuesday views read ``Hourly station entries''? \textbf{Invariant:} (I1)~Are the ``Total passengers'' values at the bottom for Monday and Tuesday 180 and 185 people respectively? (I2)~The switching between Monday and Tuesday and the correspondence of the 08:00--12:00 time points.}
\label{fig:g-charts-01-trend-L1}
\end{figure*}

\begin{figure*}[p]
\centering
\pdfbookmark[2]{Charts / Passenger-flow line chart / L2}{bm-charts__01-trend__L2}
\includegraphics[width=\linewidth]{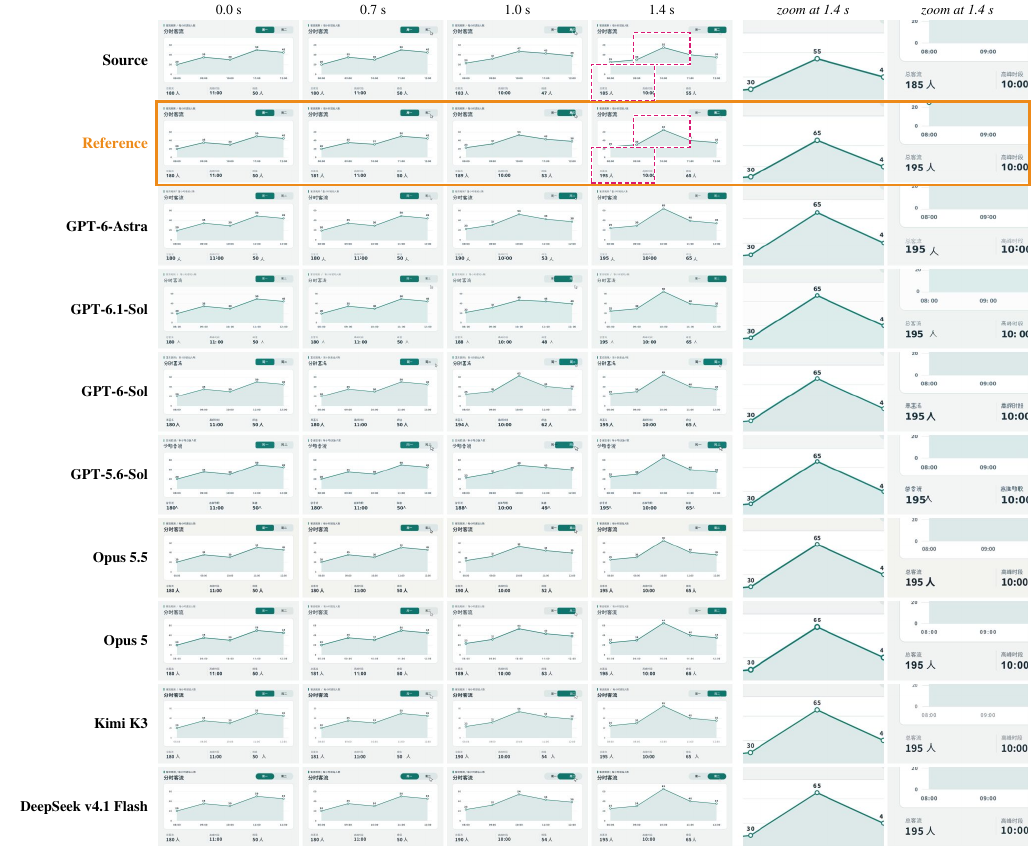}
\caption{\gallerycaptionfont \textbf{Charts / Passenger-flow line chart / L2.} \emph{Instruction:} ``Change the number of station entries at 10:00 on Tuesday to 65 people.'' \textbf{Direct:} (D1)~Is the line point at 10:00 on Tuesday labelled 65 above it? \textbf{Derived:} (R1)~Does Tuesday's line reach its highest point at 10:00, with that point above the 60-person gridline? (R2)~Is ``Total passengers'' at the bottom for Tuesday 195 people? (R3)~Is ``Peak'' at the bottom for Tuesday 65 people? \textbf{Invariant:} (I1)~Is the line point at 10:00 on Monday still labelled 30? (I2)~The switching between Monday and Tuesday and the correspondence of the 08:00--12:00 time points.}
\label{fig:g-charts-01-trend-L2}
\end{figure*}

\begin{figure*}[p]
\centering
\pdfbookmark[2]{Charts / Passenger-flow line chart / L3}{bm-charts__01-trend__L3}
\includegraphics[width=\linewidth]{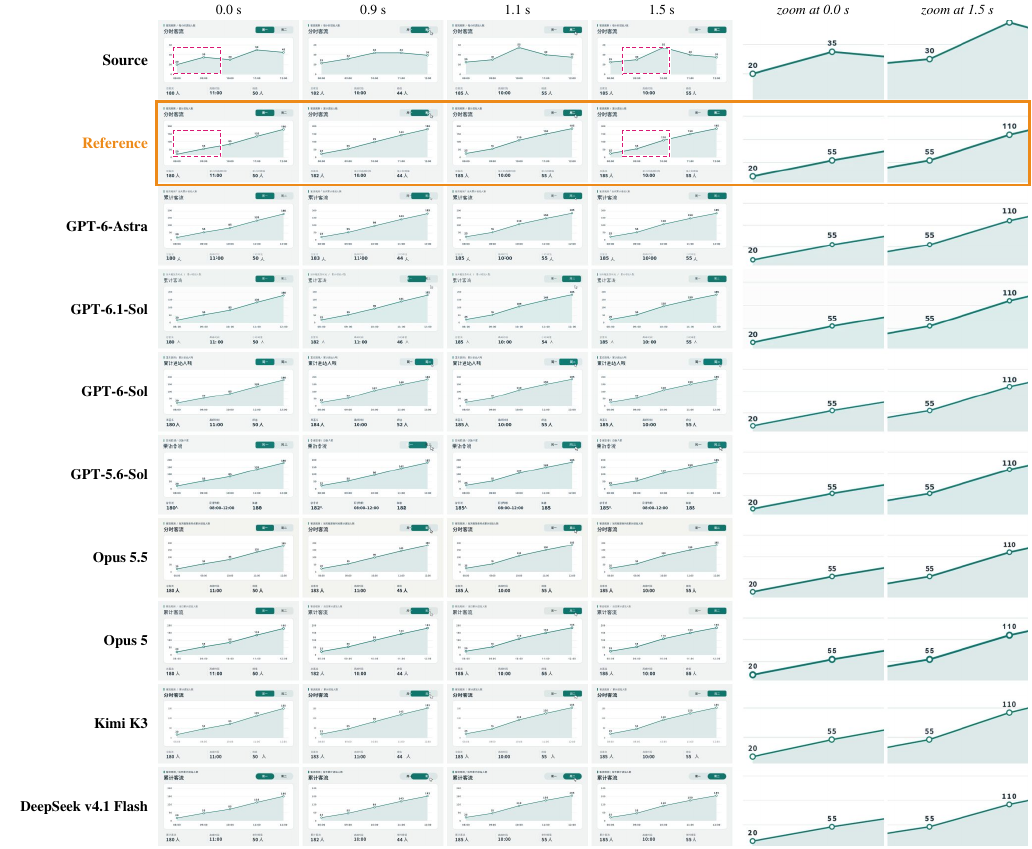}
\caption{\gallerycaptionfont \textbf{Charts / Passenger-flow line chart / L3.} \emph{Instruction:} ``Make the passenger-flow line show the cumulative number of station entries up to each time point of the day.'' \textbf{Direct:} (D1)~Are Monday's five line points from 08:00 to 12:00 labelled 20, 55, 85, 135 and 180 in order? (D2)~Are Tuesday's five line points from 08:00 to 12:00 labelled 25, 55, 110, 150 and 185 in order? \textbf{Derived:} (R1)~Do the lines of both days rise on every segment, and does the value of each 12:00 point equal that day's ``Total passengers'' at the bottom? (R2)~Are ``Single-hour peak period'' and ``Single-hour peak'' at the bottom for Monday 11:00 and 50 people respectively? (R3)~Are ``Single-hour peak period'' and ``Single-hour peak'' at the bottom for Tuesday 10:00 and 55 people respectively? \textbf{Invariant:} (I1)~The switching between Monday and Tuesday and the correspondence of the 08:00--12:00 time points.}
\label{fig:g-charts-01-trend-L3}
\end{figure*}

\begin{figure*}[p]
\centering
\pdfbookmark[2]{Charts / Desk-occupancy heatmap / L3}{bm-charts__06-heatmap__L3}
\includegraphics[width=\linewidth]{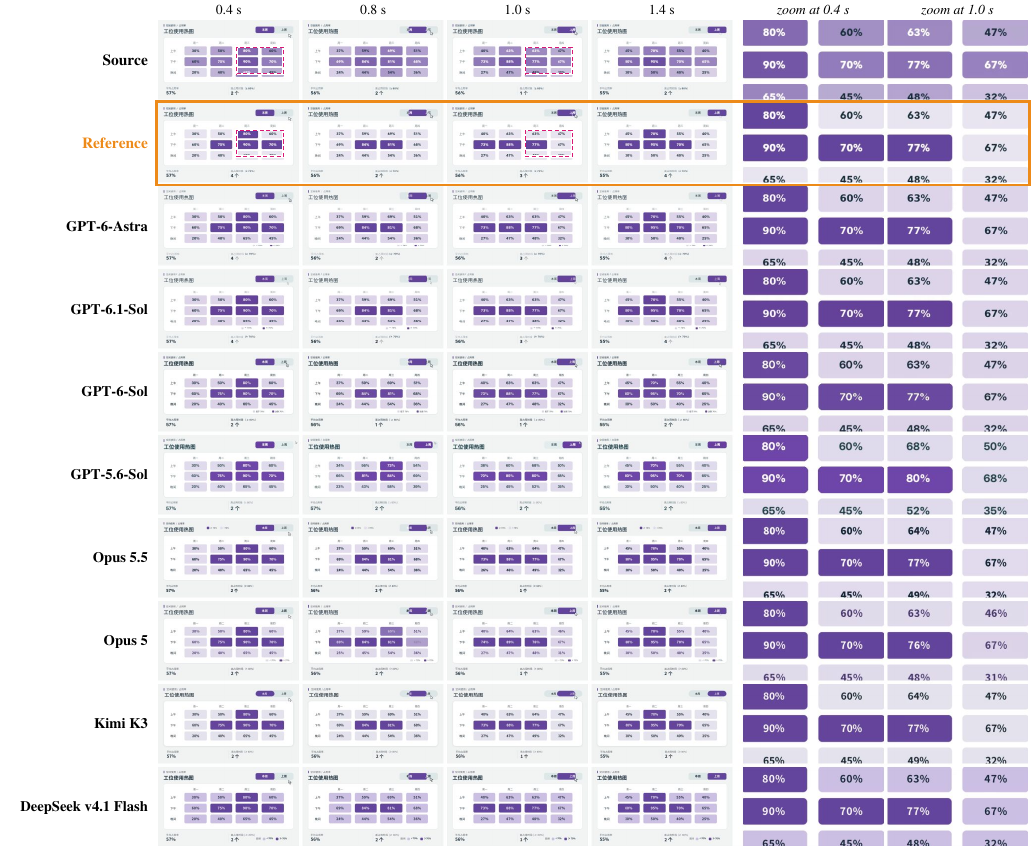}
\caption{\gallerycaptionfont \textbf{Charts / Desk-occupancy heatmap / L3.} \emph{Instruction:} ``Use two colour levels to distinguish whether desk occupancy reaches 70\%.'' \textbf{Direct:} (D1)~Do the cells of both this week and last week use only two fill colours, dark purple for 70\% and above and a light colour for below 70\%? (D2)~Do the two cells labelled 70\%, this week's Thursday afternoon and last week's Tuesday morning, both have the same dark purple background as the 80\% cells in their respective views? \textbf{Derived:} (R1)~Do the high-occupancy summaries at the bottom for both this week and last week indicate ``$\geq$70\%'' and show 4 slots? \textbf{Invariant:} (I1)~Are the average occupancy rates of this week and last week 57\% and 55\% respectively? (I2)~The row--column correspondence, percentages and switching order of this week and last week (except values involved in the instruction).}
\label{fig:g-charts-06-heatmap-L3}
\end{figure*}

\clearpage

\begin{figure*}[p]
\centering
\pdfbookmark[2]{Chemistry / Fluorine atom card / L1}{bm-chemistry__atom-f__L1}
\includegraphics[width=\linewidth]{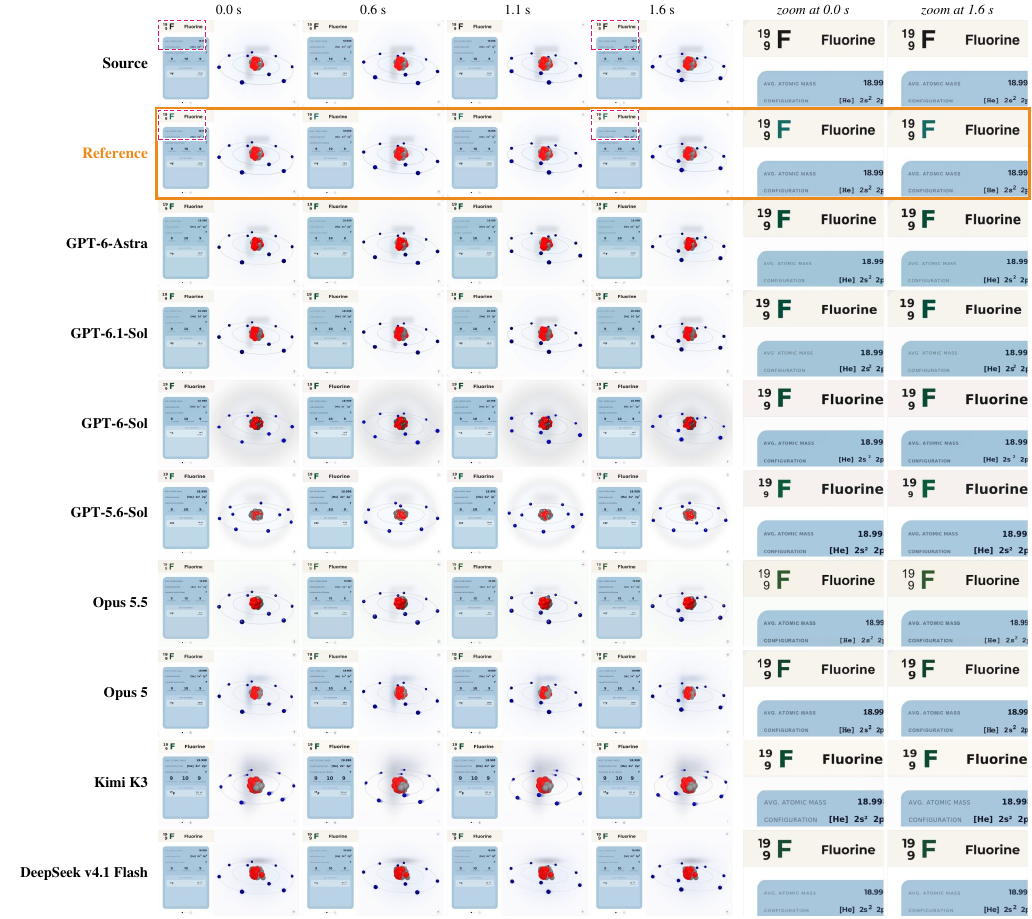}
\caption{\gallerycaptionfont \textbf{Chemistry / Fluorine atom card / L1.} \emph{Instruction:} ``Change the fluorine element symbol F at the upper left to dark green.'' \textbf{Direct:} (D1)~Is the element symbol F in the title at the upper left dark green? \textbf{Invariant:} (I1)~Are the 19 and 9 stacked vertically to the left of F both black? (I2)~Fluorine's atomic number, mass number and the particle properties not involved are kept, and the animated atom display is kept.}
\label{fig:g-chemistry-atom-f-L1}
\end{figure*}

\begin{figure*}[p]
\centering
\pdfbookmark[2]{Chemistry / Fluorine atom card / L2}{bm-chemistry__atom-f__L2}
\includegraphics[width=\linewidth]{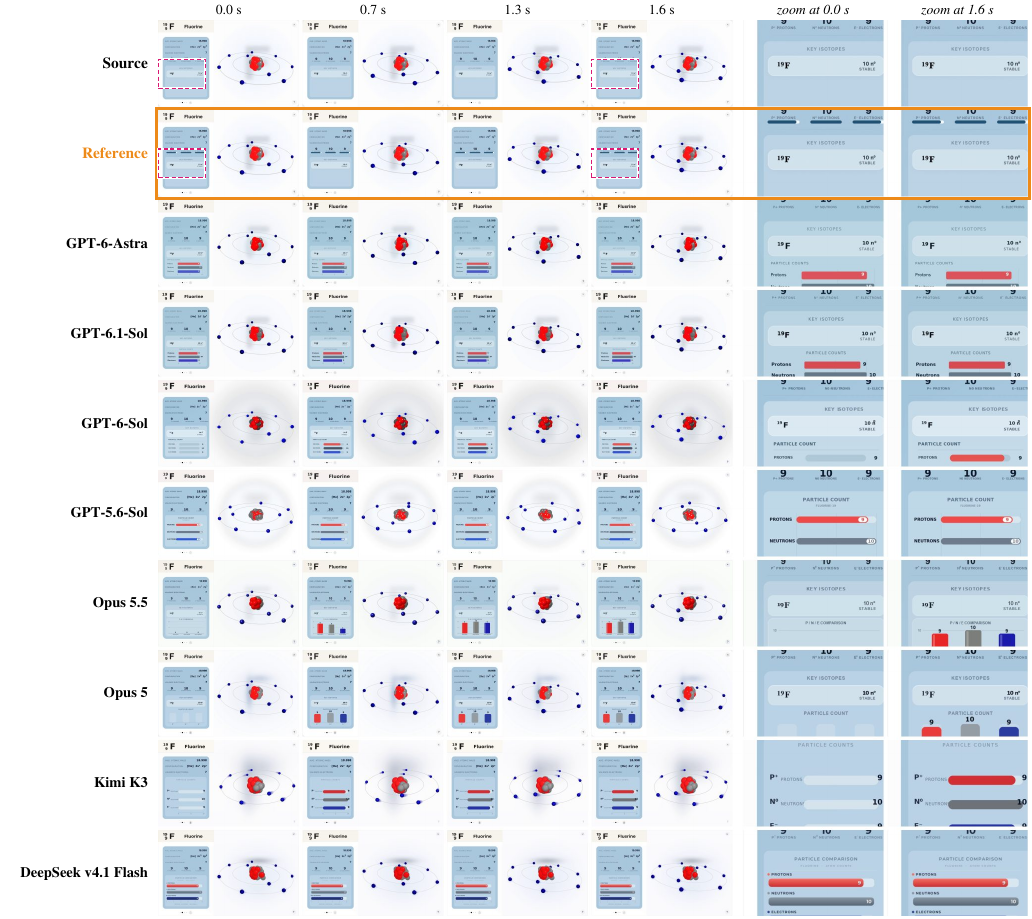}
\caption{\gallerycaptionfont \textbf{Chemistry / Fluorine atom card / L2.} \emph{Instruction:} ``Compare the numbers of protons, neutrons and electrons on the fluorine card with bar charts.'' \textbf{Direct:} (D1)~Do the three fields PROTONS, NEUTRONS and ELECTRONS in the particle-count area on the left show 9, 10 and 9 respectively, each with a quantity bar below it? \textbf{Derived:} (R1)~Are the dark filled segments of the PROTONS and ELECTRONS quantity bars equal in length? (R2)~Taking the dark segment of the NEUTRONS bar as the full length, is the dark segment of the PROTONS bar about 90\% of it (85\% to 95\% is acceptable)? \textbf{Invariant:} (I1)~Fluorine's atomic number, mass number and the particle properties not involved are kept, and the animated atom display is kept.}
\label{fig:g-chemistry-atom-f-L2}
\end{figure*}

\begin{figure*}[p]
\centering
\pdfbookmark[2]{Chemistry / Fluorine atom card / L3}{bm-chemistry__atom-f__L3}
\includegraphics[width=\linewidth]{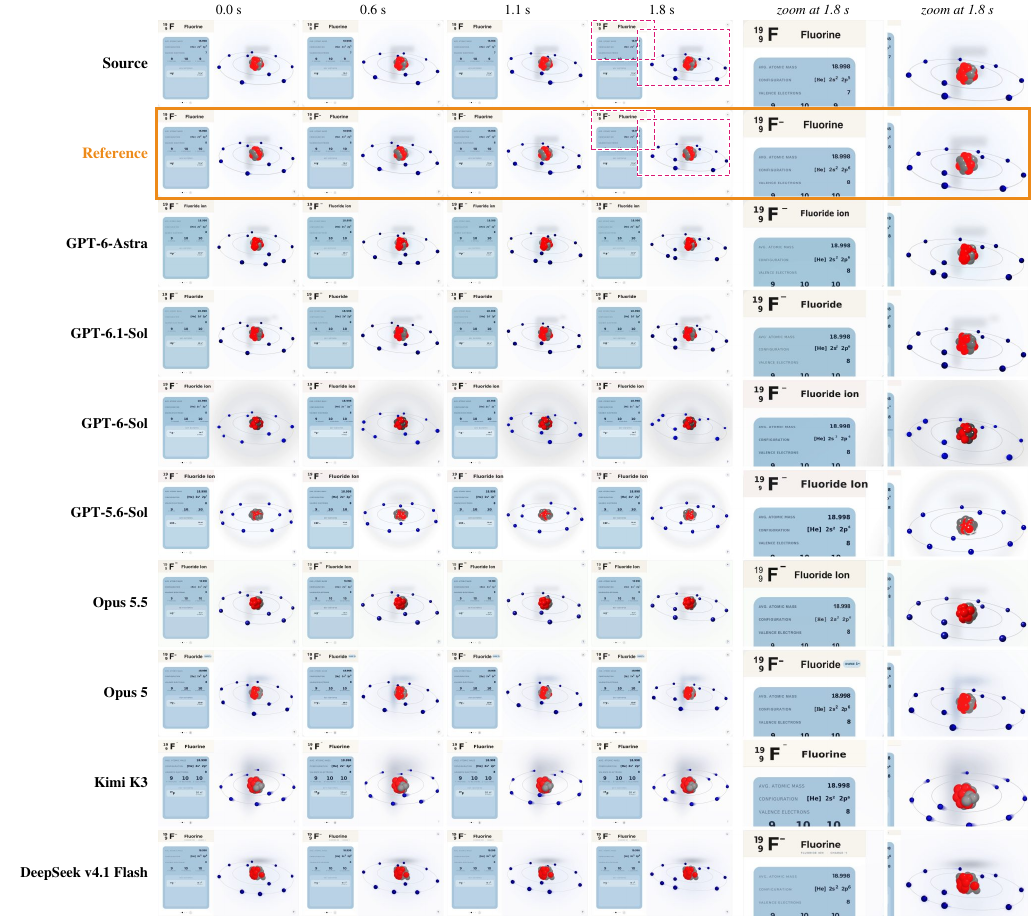}
\caption{\gallerycaptionfont \textbf{Chemistry / Fluorine atom card / L3.} \emph{Instruction:} ``Change the neutral fluorine atom into an F$^{-}$ ion.'' \textbf{Direct:} (D1)~Is the fluorine element symbol at the upper left, together with its superscript charge at the upper right, written as F$^{-}$? \textbf{Derived:} (R1)~Does the ELECTRONS field of the fluoride ion card show 10? (R2)~Does the CONFIGURATION field of F$^{-}$ on the left show [He] 2s$^{2}$ 2p$^{6}$ (it may also be written as [Ne]; when the inner shells are expanded, the corresponding full configuration is 1s$^{2}$ 2s$^{2}$ 2p$^{6}$, and the subshell order may differ)? (R3)~On the orbits of the fluoride ion model on the right, are there 2 and 8 blue electron balls respectively from the inside out? (R4)~Does the VALENCE ELECTRONS field of F$^{-}$ on the left show 8? \textbf{Invariant:} (I1)~Does the PROTONS field of the fluoride ion card show 9? (I2)~Fluorine's atomic number, mass number and the particle properties not involved are kept, and the animated atom display is kept.}
\label{fig:g-chemistry-atom-f-L3}
\end{figure*}

\begin{figure*}[p]
\centering
\pdfbookmark[2]{Chemistry / Phosphorus atom card / L3}{bm-chemistry__atom-p__L3}
\includegraphics[width=\linewidth]{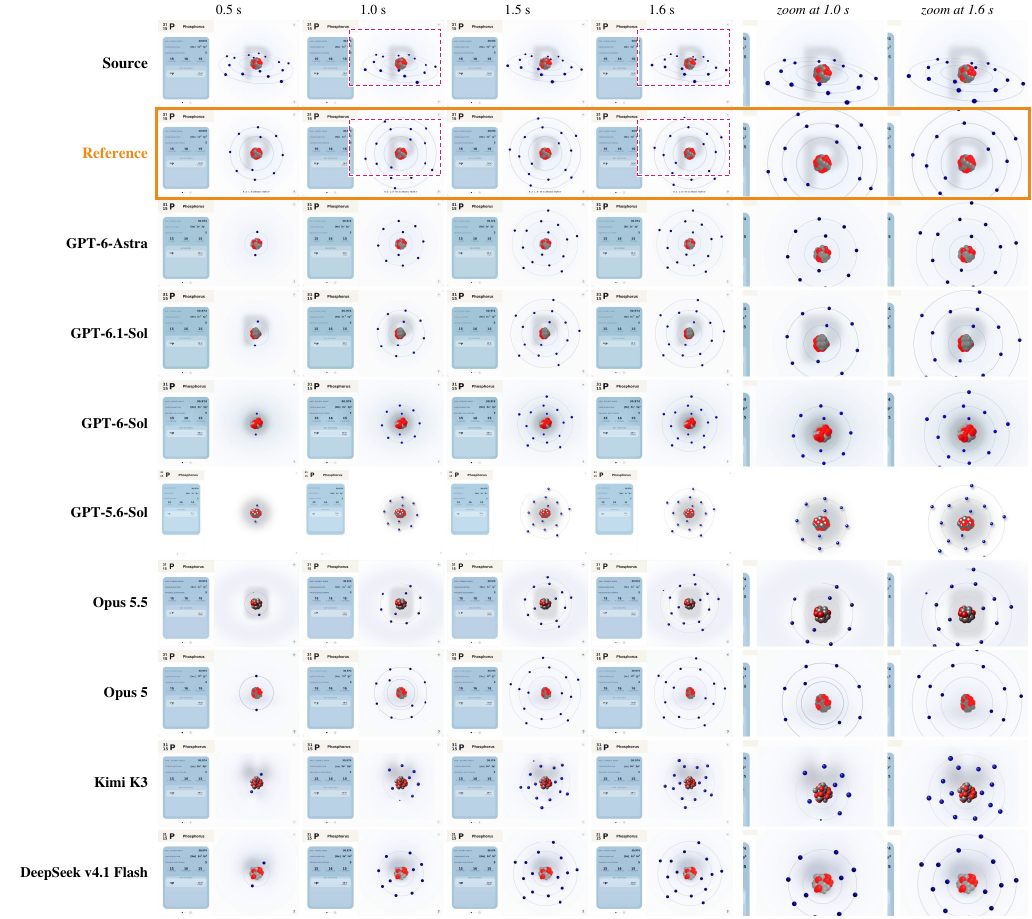}
\caption{\gallerycaptionfont \textbf{Chemistry / Phosphorus atom card / L3.} \emph{Instruction:} ``Change the phosphorus atom model on the right into a front-view animation that reveals the electron shells one by one from the inside out.'' \textbf{Direct:} (D1)~At the start of the video, does only one circular orbit appear around the phosphorus nucleus on the right, with 2 blue electron balls on it? (D2)~When the second circular orbit appears, does the inner ring keep its 2 blue balls, and are there 8 blue balls on the newly appeared outer ring? \textbf{Derived:} (R1)~At the end of the video, do the three concentric circular orbits of the model on the right carry 2, 8 and 5 blue balls respectively from the inside out? (R2)~Does the number of displayed electrons under Shown below the model on the right change to 2/15, 10/15 and 15/15 in turn as the orbits appear? (R3)~After all three shells have appeared, do the blue electron balls continue to move along their respective circular orbits? \textbf{Invariant:} (I1)~Phosphorus's atomic number and particle counts remain physically consistent, and the card content not involved is kept.}
\label{fig:g-chemistry-atom-p-L3}
\end{figure*}

\clearpage

\begin{figure*}[p]
\centering
\pdfbookmark[2]{Circuits / RC low-pass filter / L1}{bm-electric__04_rc_low_pass__L1}
\includegraphics[width=\linewidth]{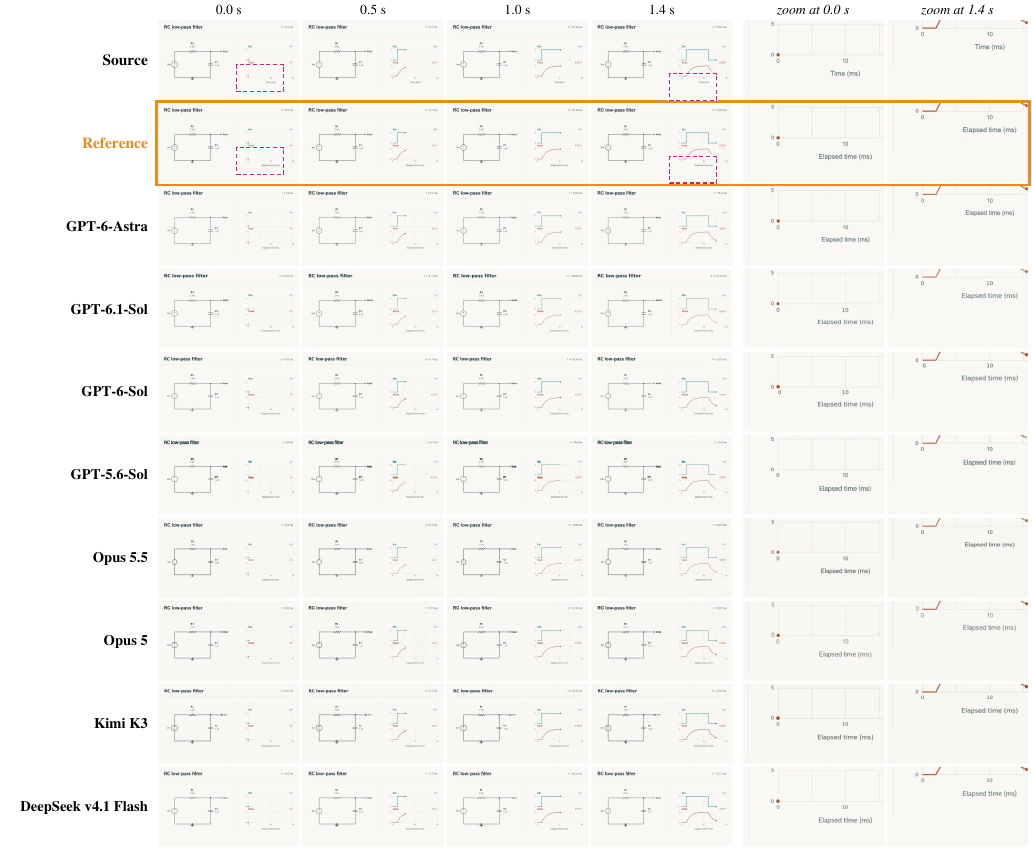}
\caption{\gallerycaptionfont \textbf{Circuits / RC low-pass filter / L1.} \emph{Instruction:} ``Change the horizontal-axis title of the low-pass filter's oscilloscope plot to ``Elapsed time (ms)''.'' \textbf{Direct:} (D1)~Does the horizontal-axis title below the waveforms on the right read ``Elapsed time (ms)''? \textbf{Invariant:} (I1)~Are the three time ticks 0, 10 and 20 marked on the horizontal axis from left to right? (I2)~The timing of the input stimulus and the voltage and time units are kept, and the circuit nodes and the two waveform traces are identifiable.}
\label{fig:g-electric-04-rc-low-pass-L1}
\end{figure*}

\begin{figure*}[p]
\centering
\pdfbookmark[2]{Circuits / RC low-pass filter / L2}{bm-electric__04_rc_low_pass__L2}
\includegraphics[width=\linewidth]{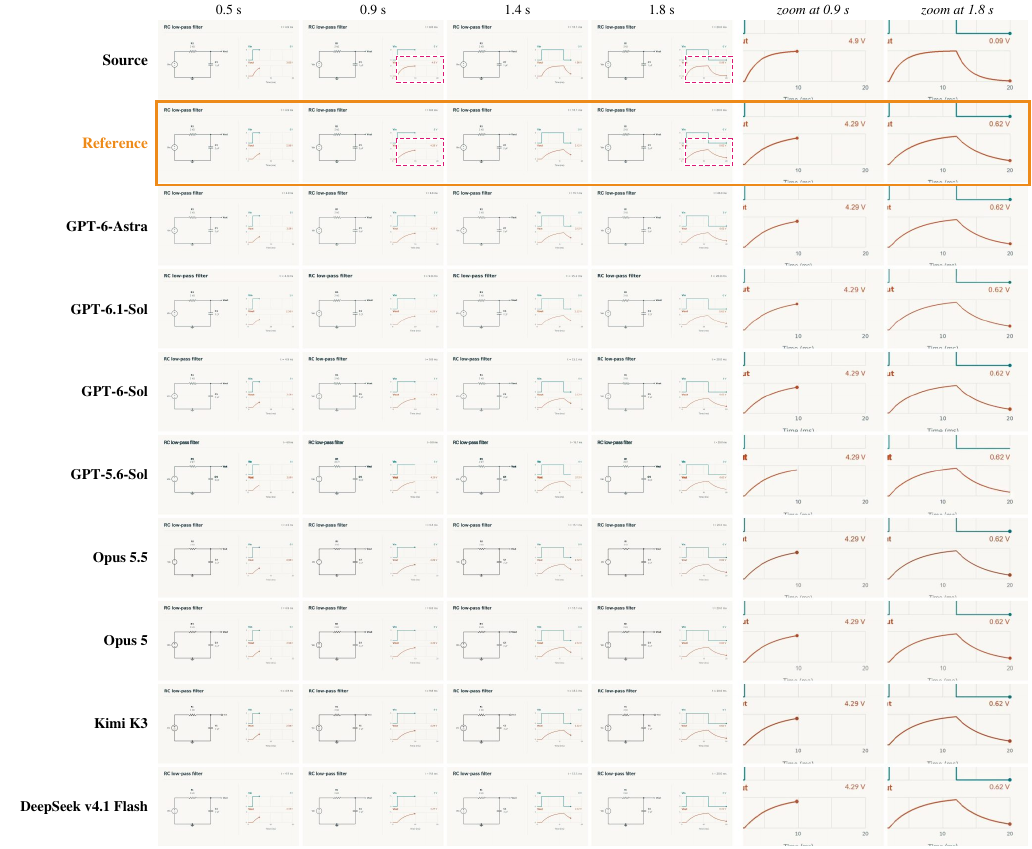}
\caption{\gallerycaptionfont \textbf{Circuits / RC low-pass filter / L2.} \emph{Instruction:} ``Change the capacitor C1 of the RC low-pass filter to 2 $\mu$F.'' \textbf{Direct:} (D1)~Is C1, connected from Vout to ground, labelled 2 $\mu$F? \textbf{Derived:} (R1)~When the high plateau of Vin ends at about 12 ms, does Vout reach a peak of about 4.6 V? (R2)~After Vin drops back to zero, does Vout fall smoothly from its peak to about 0.62 V at 20 ms? \textbf{Invariant:} (I1)~Is R1, between Vin and Vout, labelled 2 k$\Omega$? (I2)~The timing of the input stimulus and the voltage and time units are kept, and the circuit nodes and the two waveform traces are identifiable.}
\label{fig:g-electric-04-rc-low-pass-L2}
\end{figure*}

\begin{figure*}[p]
\centering
\pdfbookmark[2]{Circuits / RC low-pass filter / L3}{bm-electric__04_rc_low_pass__L3}
\includegraphics[width=\linewidth]{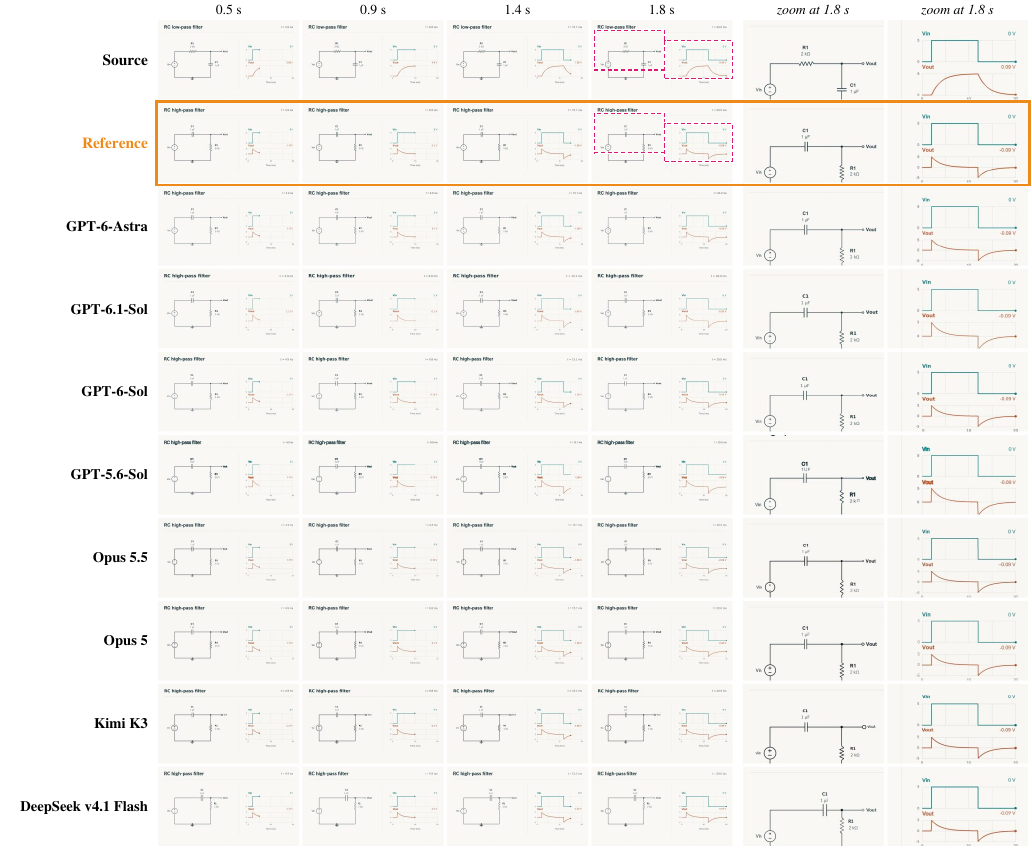}
\caption{\gallerycaptionfont \textbf{Circuits / RC low-pass filter / L3.} \emph{Instruction:} ``Turn the RC low-pass filter into a high-pass filter using the original circuit's R1 and C1.'' \textbf{Direct:} (D1)~Is C1, labelled 1 $\mu$F, connected in series in the upper line from Vin to Vout? (D2)~Is R1, labelled 2 k$\Omega$, connected between Vout and ground? \textbf{Derived:} (R1)~Does the title at the upper left state that this is an RC high-pass filter? (R2)~At about 2 ms, when Vin rises from 0 V to 5 V, does Vout first jump to about +5 V and then decay towards 0 V? (R3)~At about 12 ms, when Vin drops from 5 V to 0 V, does Vout first jump to about $-$5 V, then recover from the negative side and approach 0 V at the end of the video? \textbf{Invariant:} (I1)~The timing of the input stimulus and the voltage and time units are kept, and the circuit nodes and the two waveform traces are identifiable.}
\label{fig:g-electric-04-rc-low-pass-L3}
\end{figure*}

\begin{figure*}[p]
\centering
\pdfbookmark[2]{Circuits / Full adder / L3}{bm-electric__21_full_adder__L3}
\includegraphics[width=\linewidth]{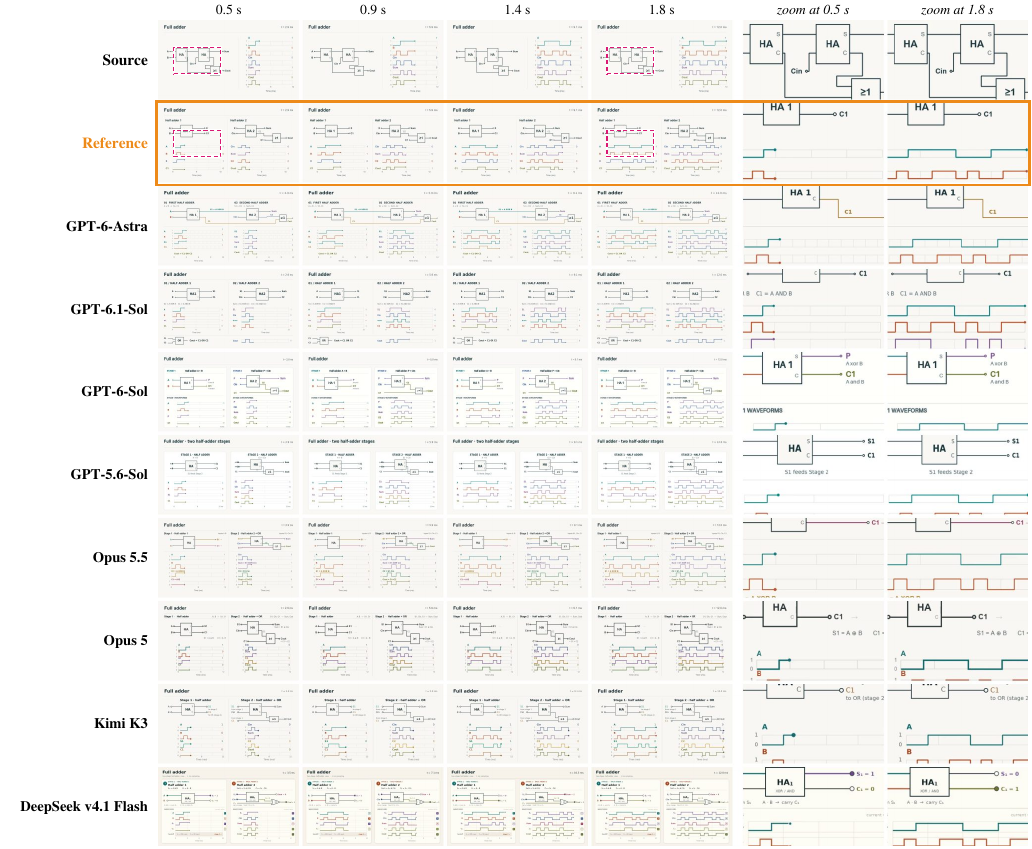}
\caption{\gallerycaptionfont \textbf{Circuits / Full adder / L3.} \emph{Instruction:} ``Show the circuit and waveforms of the full adder in separate columns for its two half-adder stages.'' \textbf{Direct:} (D1)~Does the left column, containing the first-stage half adder HA 1, show the four traces A, B, X and C1 together? (D2)~Does the right column, containing the second-stage half adder HA 2, show the Cin, Sum, C2 and final Cout traces together? \textbf{Derived:} (R1)~Viewing the 0 to 12 ms scales of the two columns in alignment, is X in the left column high when A and B differ, and is C1 high only when A and B are both high? (R2)~Viewed on the same time scale for the left and right columns, is Sum in the right column high when exactly one of X and Cin is high, and is Sum low when both are high or both are low? (R3)~Is C2 in the right column high only when X in the left column and Cin in the right column are both high? (R4)~Are the input branches of the ``$\geq$1'' gate in the right column labelled C1 and C2, and is Cout high exactly in the intervals where C1 in the left column or C2 in the right column is high? \textbf{Invariant:} (I1)~The three inputs A, B and Cin, the two outputs Sum and Cout, and the digital timing demonstration are kept.}
\label{fig:g-electric-21-full-adder-L3}
\end{figure*}

\clearpage

\begin{figure*}[p]
\centering
\pdfbookmark[2]{Mathematics / How variance adds / L1}{bm-math__056_HowVarianceAdds__L1}
\includegraphics[width=\linewidth]{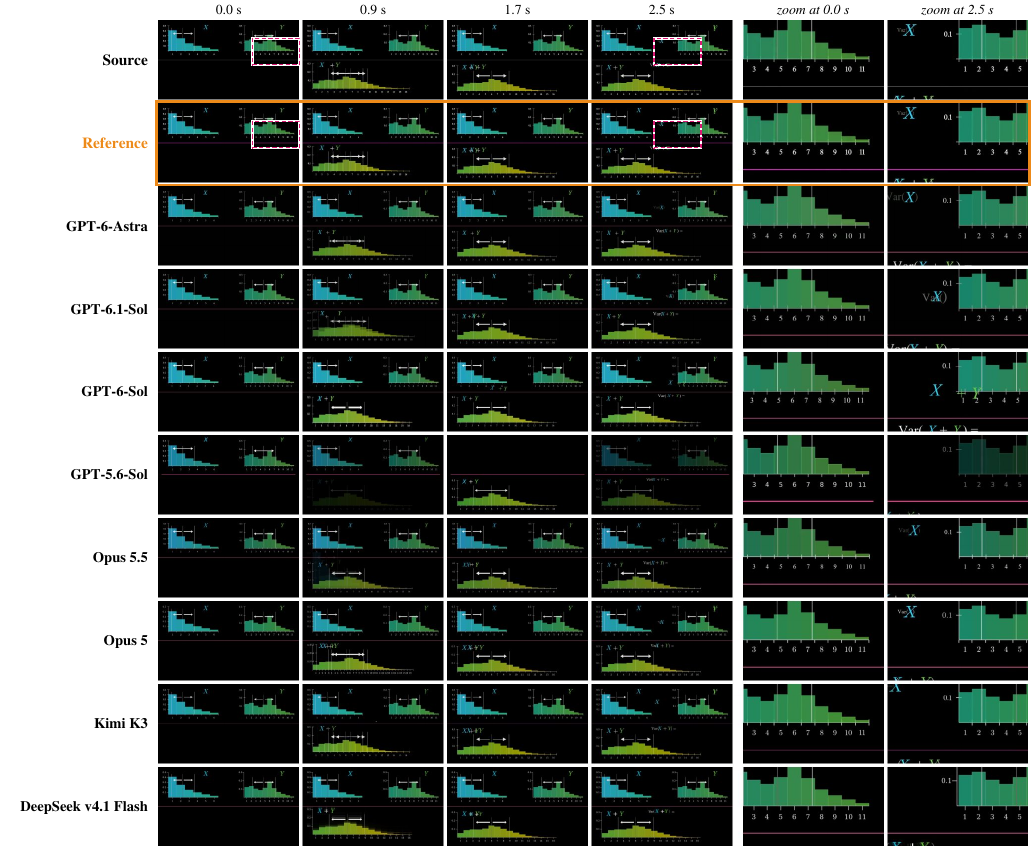}
\caption{\gallerycaptionfont \textbf{Mathematics / How variance adds / L1.} \emph{Instruction:} ``Change the dividing line between the upper and lower distribution panels to pink.'' \textbf{Direct:} (D1)~Is the horizontal line separating the upper row of input distributions from the lower row with the sum distribution pink? \textbf{Invariant:} (I1)~In the upper row, are the bars of the X distribution teal and the bars of the Y distribution green? (I2)~The identities of the three distributions X, Y and X+Y and their summation relation are kept.}
\label{fig:g-math-056-HowVarianceAdds-L1}
\end{figure*}

\begin{figure*}[p]
\centering
\pdfbookmark[2]{Mathematics / How variance adds / L2}{bm-math__056_HowVarianceAdds__L2}
\includegraphics[width=\linewidth]{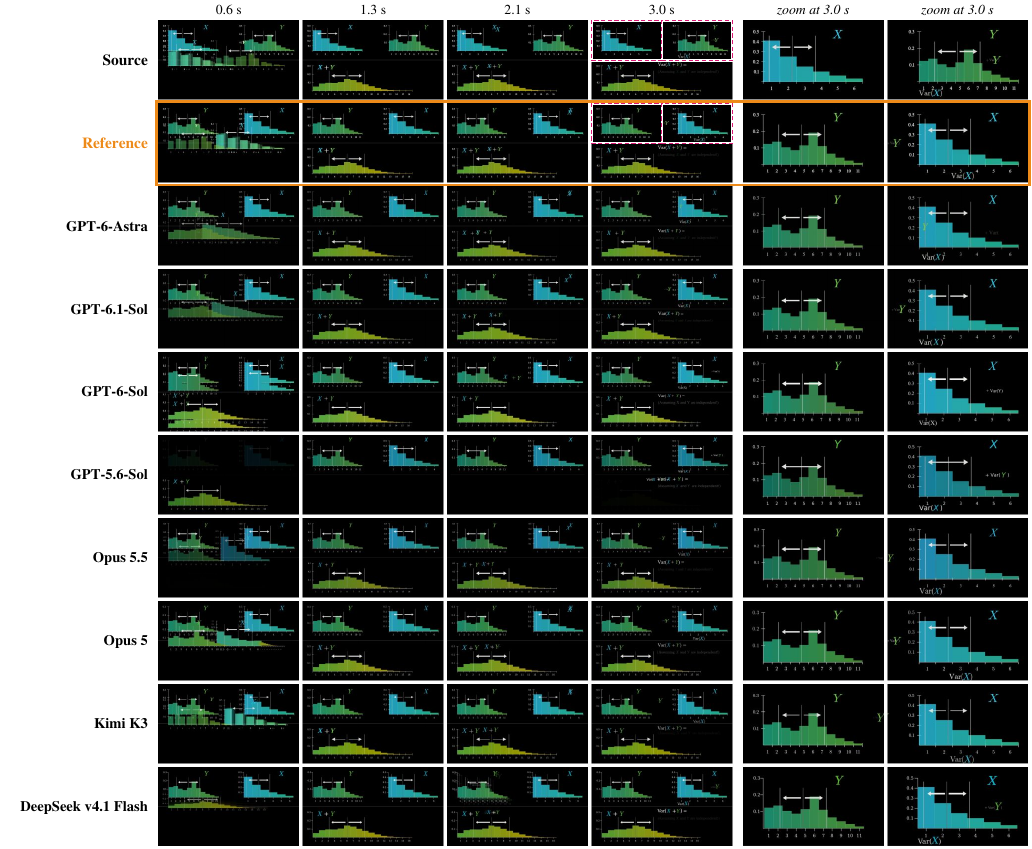}
\caption{\gallerycaptionfont \textbf{Mathematics / How variance adds / L2.} \emph{Instruction:} ``Swap the positions of the two input distribution plots X and Y in the upper row.'' \textbf{Direct:} (D1)~Is the upper left an eleven-bar green distribution labelled Y? (D2)~Is the upper right a six-bar decreasing teal distribution labelled X? \textbf{Derived:} (R1)~At the start of the video, above the green group of Y bars on the left and above the teal group of X bars on the right, is there in each case a pair of white arrows centred on that group's mean marker line and pointing left and right respectively? \textbf{Invariant:} (I1)~The identities of the three distributions X, Y and X+Y and their summation relation are kept.}
\label{fig:g-math-056-HowVarianceAdds-L2}
\end{figure*}

\begin{figure*}[p]
\centering
\pdfbookmark[2]{Mathematics / How variance adds / L3}{bm-math__056_HowVarianceAdds__L3}
\includegraphics[width=\linewidth]{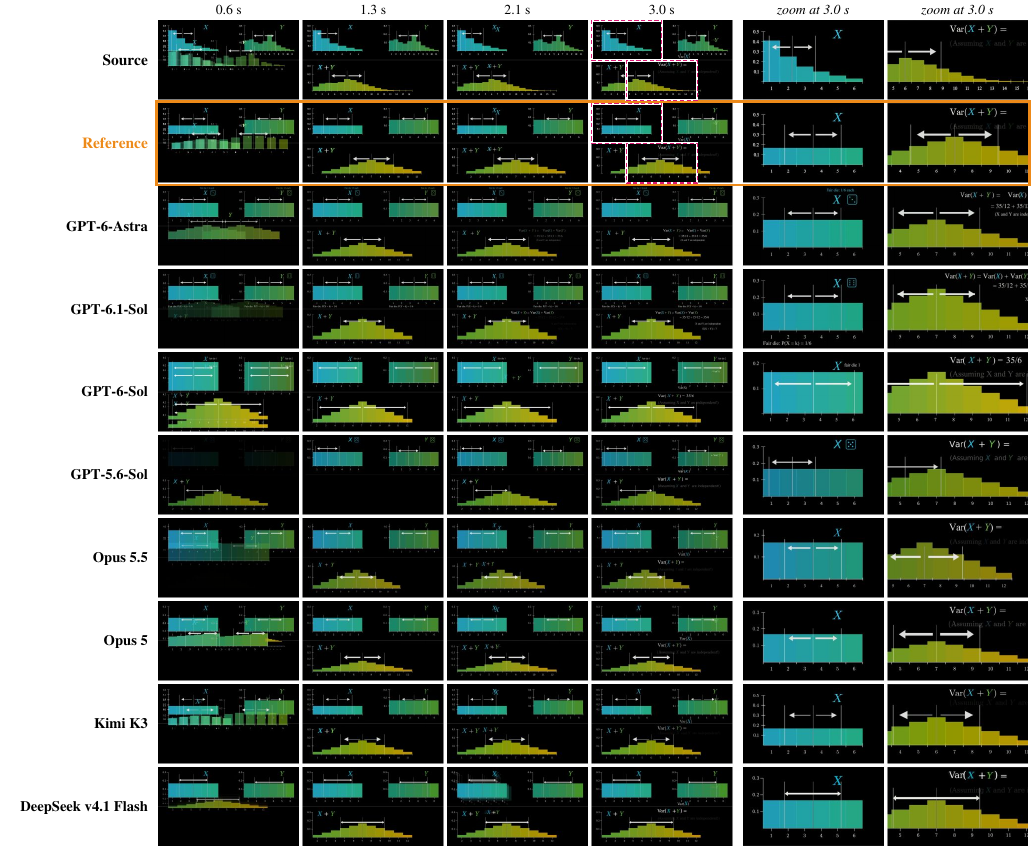}
\caption{\gallerycaptionfont \textbf{Mathematics / How variance adds / L3.} \emph{Instruction:} ``Replace X and Y by two independent fair six-sided dice.'' \textbf{Direct:} (D1)~Do the two plots X and Y in the upper row each have six bars of equal height, with horizontal coordinates from 1 to 6 in both? \textbf{Derived:} (R1)~In the X + Y plot below, are the horizontal coordinates of the first and last non-zero bars 2 and 12 respectively? (R2)~Are the X + Y bars below symmetric about 7, reaching their unique highest point at 7? (R3)~In the X + Y plot below, is the bar height at horizontal coordinate 7 about six times that at 2? (R4)~In the X + Y plot below, is the mean marker line between the two white arrows aligned with the centre of the tallest bar? \textbf{Invariant:} (I1)~The identities of the three distributions X, Y and X+Y and their summation relation are kept.}
\label{fig:g-math-056-HowVarianceAdds-L3}
\end{figure*}

\begin{figure*}[p]
\centering
\pdfbookmark[2]{Mathematics / Linear regression / L2}{bm-math__033_LinearRegression__L2}
\includegraphics[width=\linewidth]{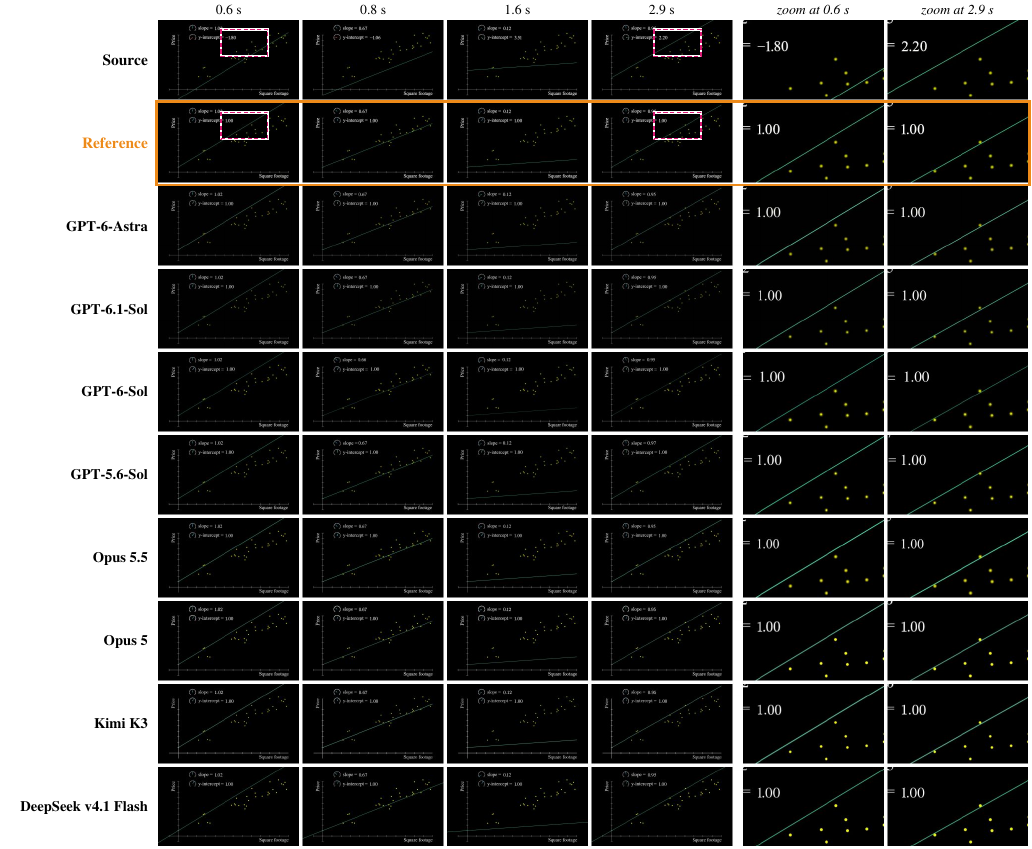}
\caption{\gallerycaptionfont \textbf{Mathematics / Linear regression / L2.} \emph{Instruction:} ``Fix the y-intercept of the regression line in the scatter plot at 1.'' \textbf{Direct:} (D1)~As the slope readout changes, does the intersection of the regression line with the vertical axis stay fixed at the first tick above the horizontal axis? \textbf{Derived:} (R1)~Does the pointer of the dial labelled y-intercept keep the same direction while the slope changes? (R2)~Does the value beside y-intercept read 1.00 throughout? \textbf{Invariant:} (I1)~The correspondence among the axes, the scatter points, the line and the slope/intercept parameter controls.}
\label{fig:g-math-033-LinearRegression-L2}
\end{figure*}

\clearpage

\begin{figure*}[p]
\centering
\pdfbookmark[2]{Physics / Three-gear chain / L1}{bm-physics__three_gear_chain__L1}
\includegraphics[width=\linewidth]{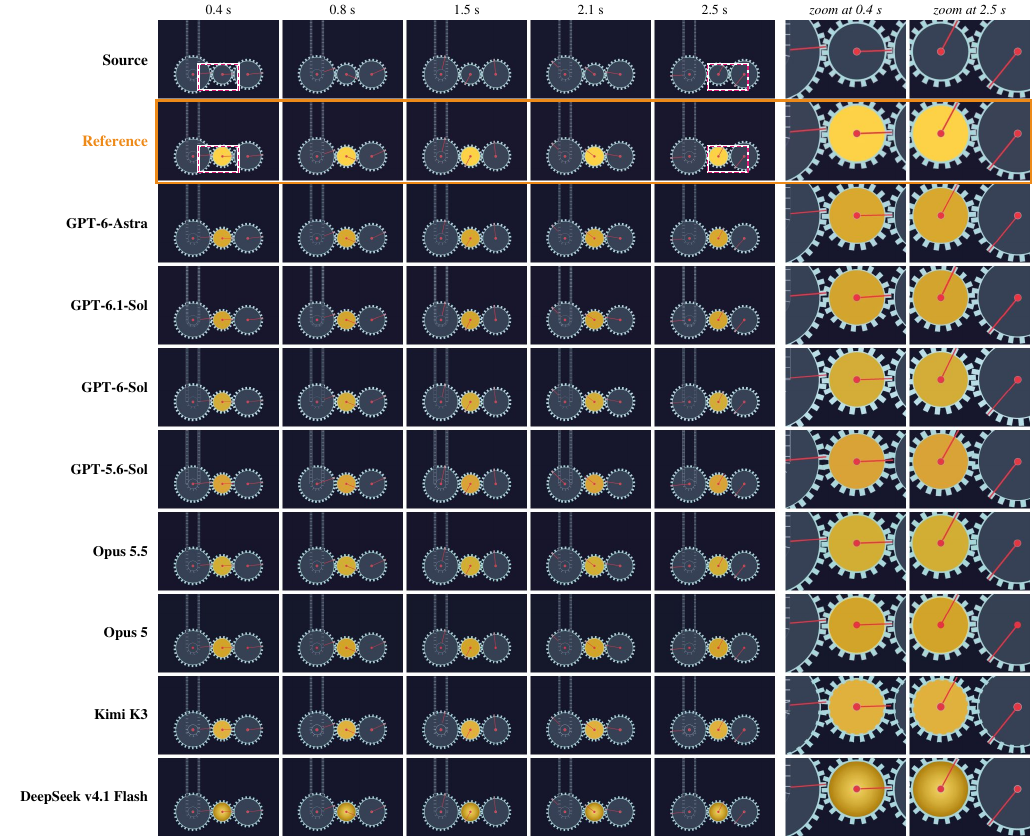}
\caption{\gallerycaptionfont \textbf{Physics / Three-gear chain / L1.} \emph{Instruction:} ``Paint the disc of the middle gear gold.'' \textbf{Direct:} (D1)~Is the disc of the middle gear gold? \textbf{Invariant:} (I1)~While the disc of the middle gear is gold, are the discs of the left and right gears both dark grey? (I2)~Does the red pointer on the gold disc rotate with the middle gear? (I3)~The left gear with the belt, the three distinguishable gears and the pointers moving with the gears are kept.}
\label{fig:g-physics-three-gear-chain-L1}
\end{figure*}

\begin{figure*}[p]
\centering
\pdfbookmark[2]{Physics / Three-gear chain / L2}{bm-physics__three_gear_chain__L2}
\includegraphics[width=\linewidth]{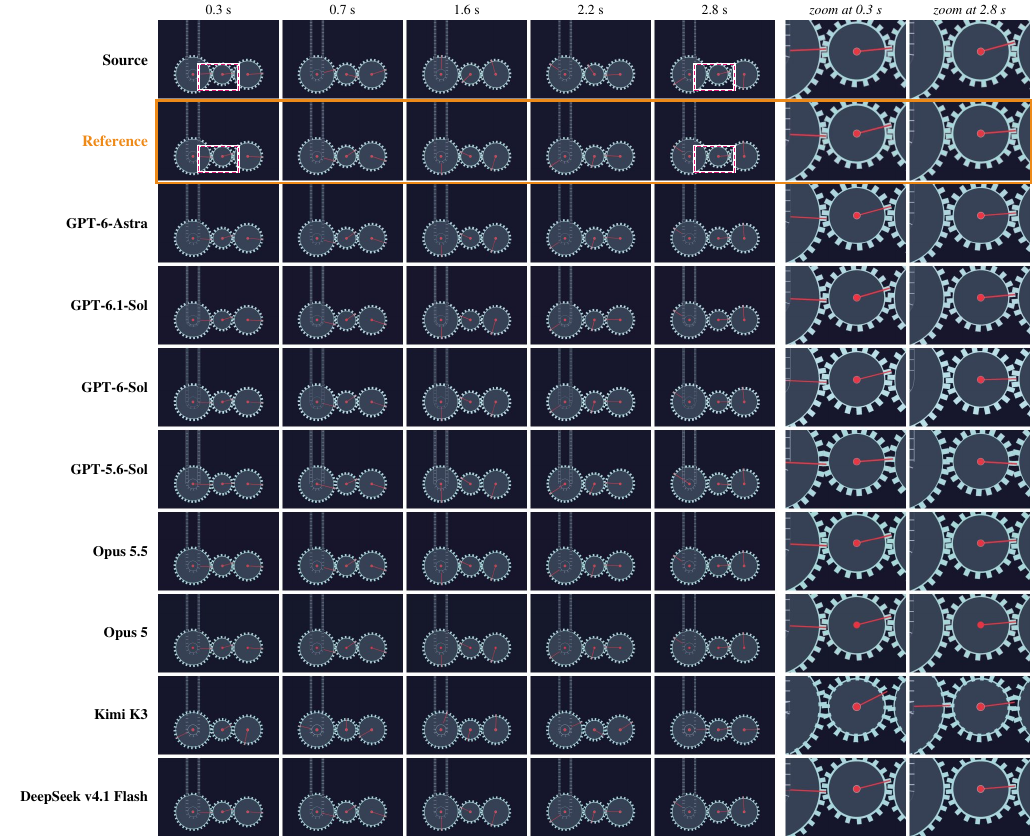}
\caption{\gallerycaptionfont \textbf{Physics / Three-gear chain / L2.} \emph{Instruction:} ``Let the belt drive the leftmost gear clockwise.'' \textbf{Direct:} (D1)~Does the red pointer of the leftmost gear turn clockwise? \textbf{Derived:} (R1)~While the left gear turns clockwise, does the red pointer on the middle gear turn anticlockwise? (R2)~While the left gear turns clockwise, does the red pointer on the rightmost gear also turn clockwise? \textbf{Invariant:} (I1)~While the three gears turn, do the teeth of the left and middle gears, and of the middle and right gears, each stay in mesh? (I2)~The left gear with the belt, the three distinguishable gears and the pointers moving with the gears are kept.}
\label{fig:g-physics-three-gear-chain-L2}
\end{figure*}

\begin{figure*}[p]
\centering
\pdfbookmark[2]{Physics / Three-gear chain / L3}{bm-physics__three_gear_chain__L3}
\includegraphics[width=\linewidth]{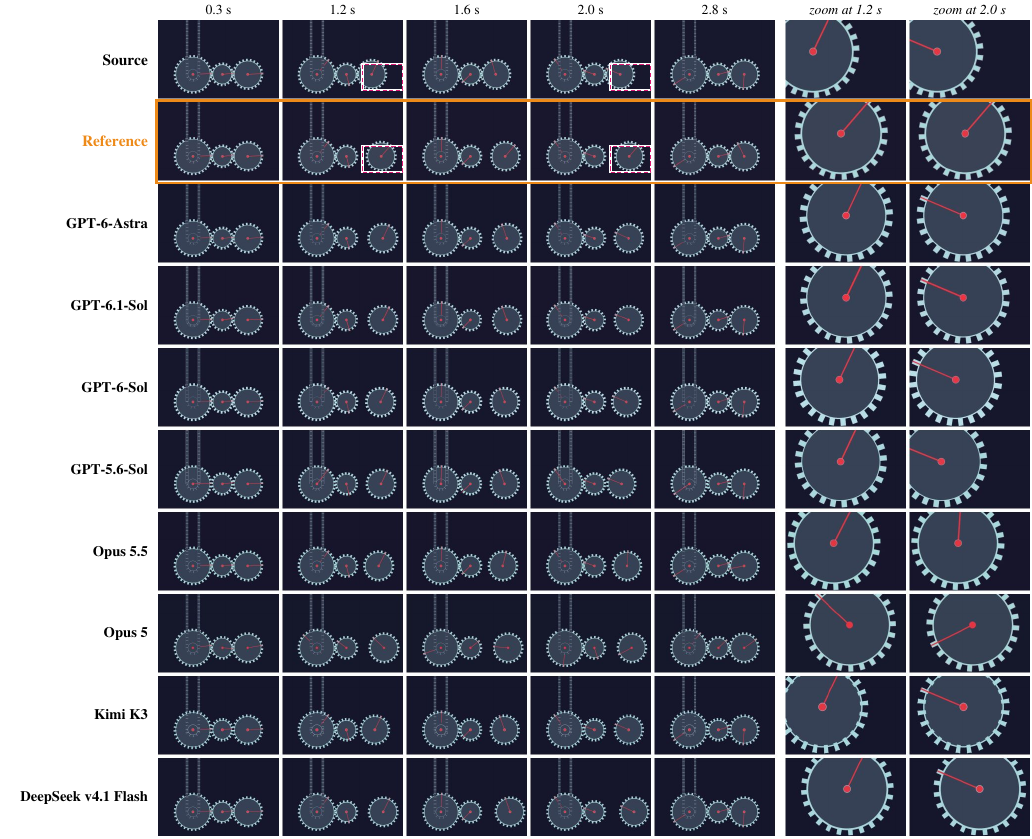}
\caption{\gallerycaptionfont \textbf{Physics / Three-gear chain / L3.} \emph{Instruction:} ``Make the rightmost gear slide out of mesh to the right while the gears turn, then slide back into place.'' \textbf{Direct:} (D1)~After the three gears start turning, does the axis of the rightmost gear move away to the right until a clear gap appears between it and the tooth tips of the middle gear? (D2)~Does the rightmost gear then slide back to the left and mesh with the middle gear again? \textbf{Derived:} (R1)~Does the rightmost gear's red pointer stay at one orientation for a while when the middle and right gears are apart? (R2)~After re-meshing, does the rightmost gear's red pointer resume turning, in the direction opposite to the middle gear? \textbf{Invariant:} (I1)~At the opening, do the left gear with the belt, the middle gear and the right gear mesh with one another in sequence? (I2)~While the middle and right gears are apart, do the red pointers of the left gear and the middle gear keep turning in opposite directions? (I3)~The left gear with the belt, the three distinguishable gears and the pointers moving with the gears are kept.}
\label{fig:g-physics-three-gear-chain-L3}
\end{figure*}

\begin{figure*}[p]
\centering
\pdfbookmark[2]{Physics / U-tube on a cart / L3}{bm-physics__u_tube_fluid_oscillation__L3}
\includegraphics[width=\linewidth]{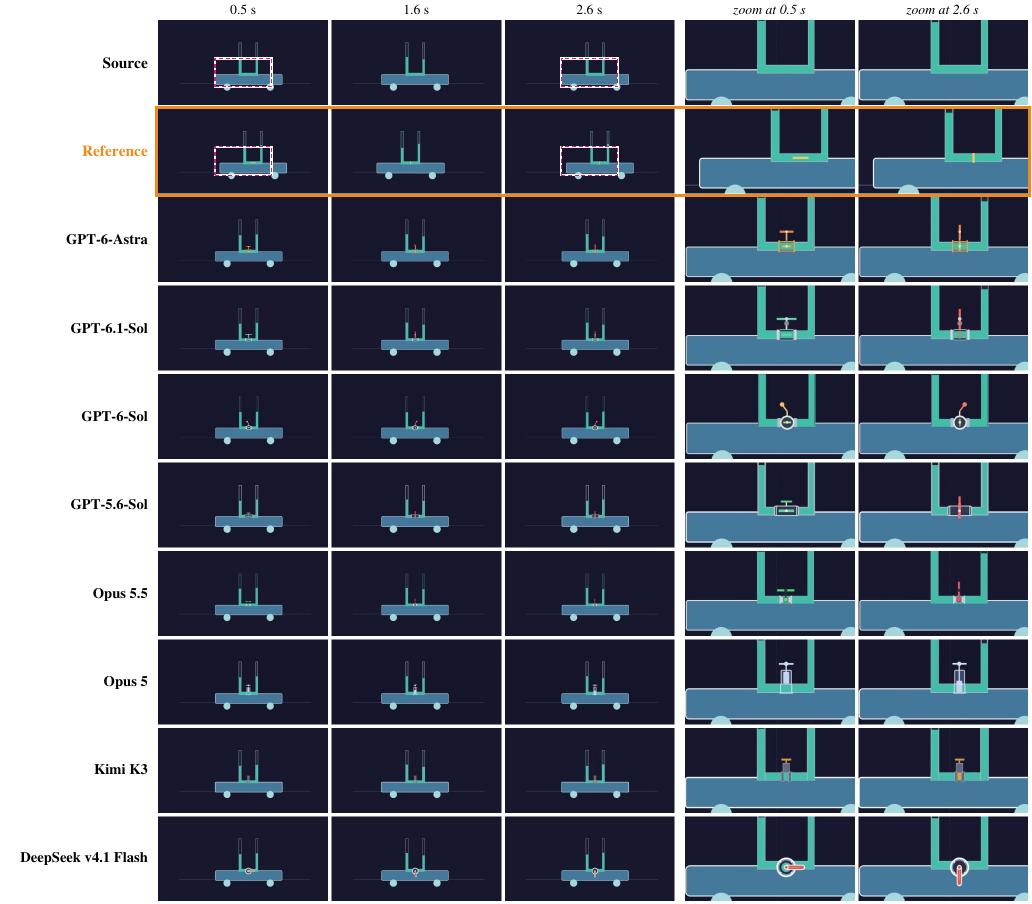}
\caption{\gallerycaptionfont \textbf{Physics / U-tube on a cart / L3.} \emph{Instruction:} ``Install a valve at the bottom of the U-tube and have it close while the cart is in motion.'' \textbf{Direct:} (D1)~Is there a gold valve handle at the centre of the horizontal tube at the bottom of the U-tube? \textbf{Derived:} (R1)~Does the gold valve handle at the centre of the bottom horizontal tube turn from horizontal to vertical? (R2)~After the gold valve handle turns vertical, does the distance from the left liquid surface to the left tube mouth remain unchanged? (R3)~After the gold valve handle turns vertical, does the distance from the right liquid surface to the right tube mouth remain unchanged? \textbf{Invariant:} (I1)~While the gold valve handle is still horizontal, can the liquid surface in the left arm be seen falling while that in the right arm rises? (I2)~After the gold valve handle is upright and the liquid surfaces on both sides have settled, does the blue cart still move along the horizontal track? (I3)~The cart's motion along the track, the U-tube structure and the consistent total liquid volume in the two arms are kept.}
\label{fig:g-physics-u-tube-fluid-oscillation-L3}
\end{figure*}

\clearpage

\begin{figure*}[p]
\centering
\pdfbookmark[2]{Web interfaces / Salon calendar / L1}{bm-web__15-salon-calendar__L1}
\includegraphics[width=\linewidth]{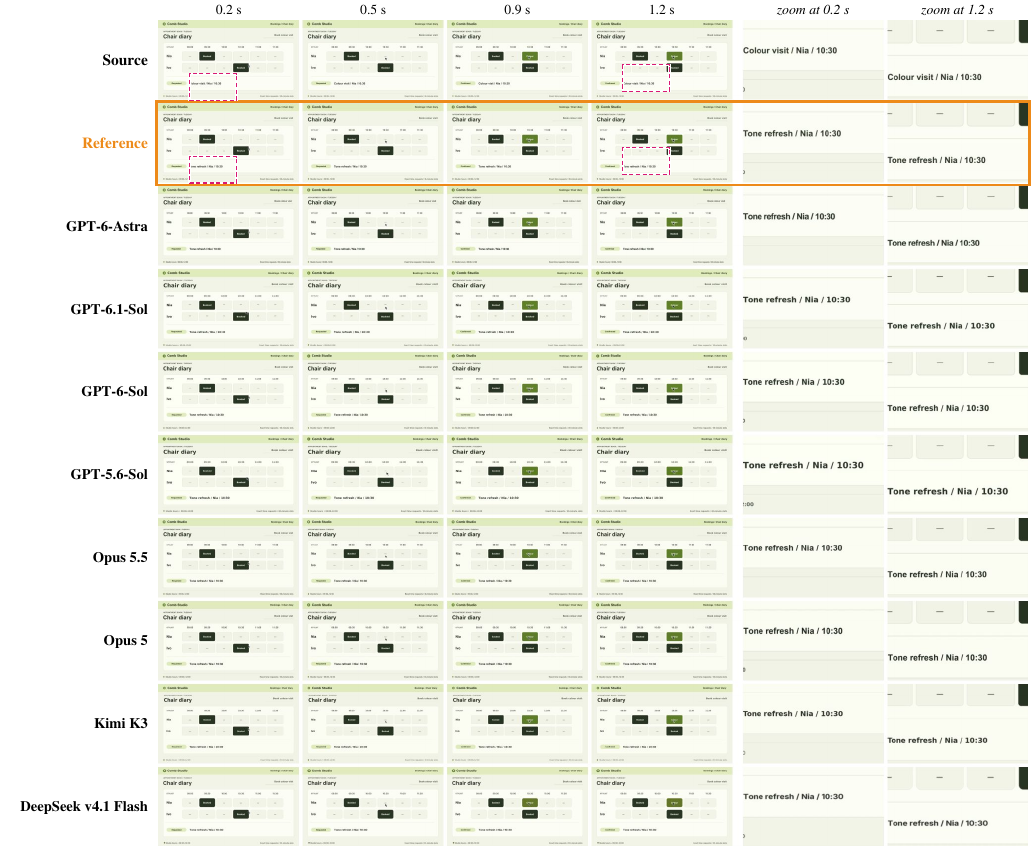}
\caption{\gallerycaptionfont \textbf{Web interfaces / Salon calendar / L1.} \emph{Instruction:} ``Change Colour visit in the booking summary at the bottom of the schedule to Tone refresh.'' \textbf{Direct:} (D1)~Is the service name shown alongside Nia / 10:30 at the bottom of the schedule Tone refresh? \textbf{Invariant:} (I1)~The two staff members, the time-slot grid and the interaction of sending a booking request are kept.}
\label{fig:g-web-15-salon-calendar-L1}
\end{figure*}

\begin{figure*}[p]
\centering
\pdfbookmark[2]{Web interfaces / Salon calendar / L2}{bm-web__15-salon-calendar__L2}
\includegraphics[width=\linewidth]{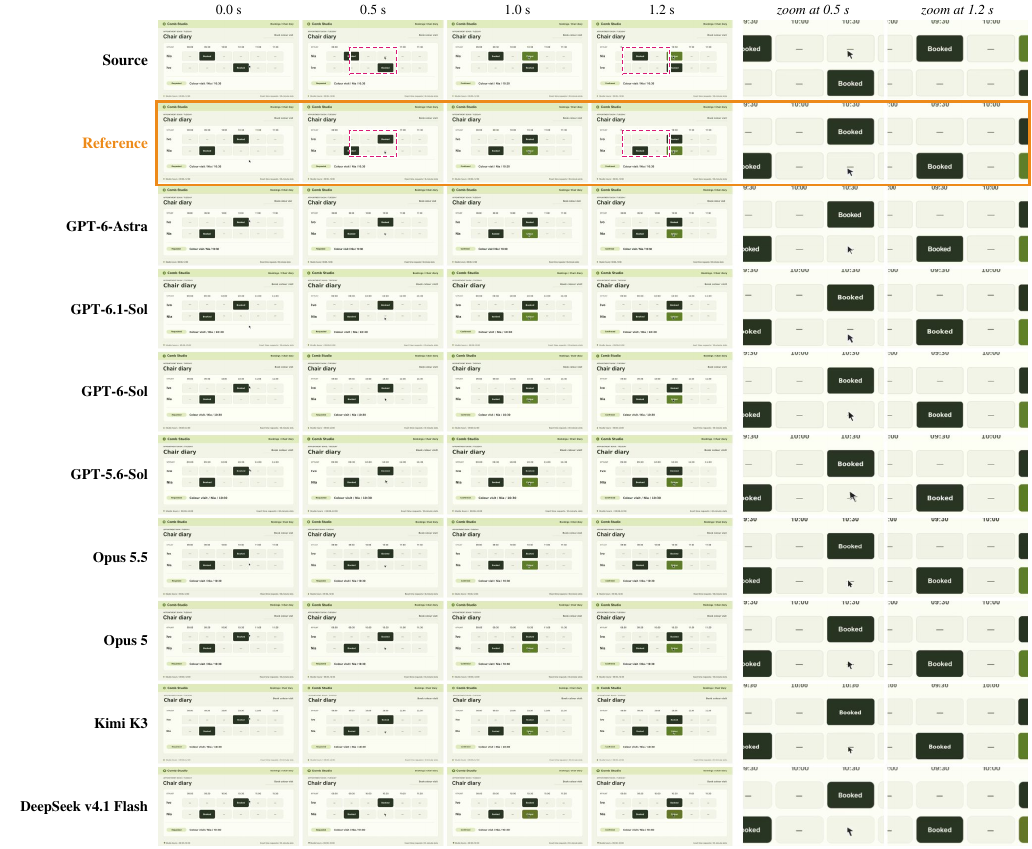}
\caption{\gallerycaptionfont \textbf{Web interfaces / Salon calendar / L2.} \emph{Instruction:} ``Move Ivo's schedule row above Nia's.'' \textbf{Direct:} (D1)~Is the topmost staff row of the schedule Ivo? \textbf{Derived:} (R1)~After booking, does the 10:30 cell of Nia in the second row show Colour? \textbf{Invariant:} (I1)~Is the 10:30 cell of Ivo in the first row still Booked? (I2)~The two staff members, the time-slot grid and the interaction of sending a booking request are kept.}
\label{fig:g-web-15-salon-calendar-L2}
\end{figure*}

\begin{figure*}[p]
\centering
\pdfbookmark[2]{Web interfaces / Salon calendar / L3}{bm-web__15-salon-calendar__L3}
\includegraphics[width=\linewidth]{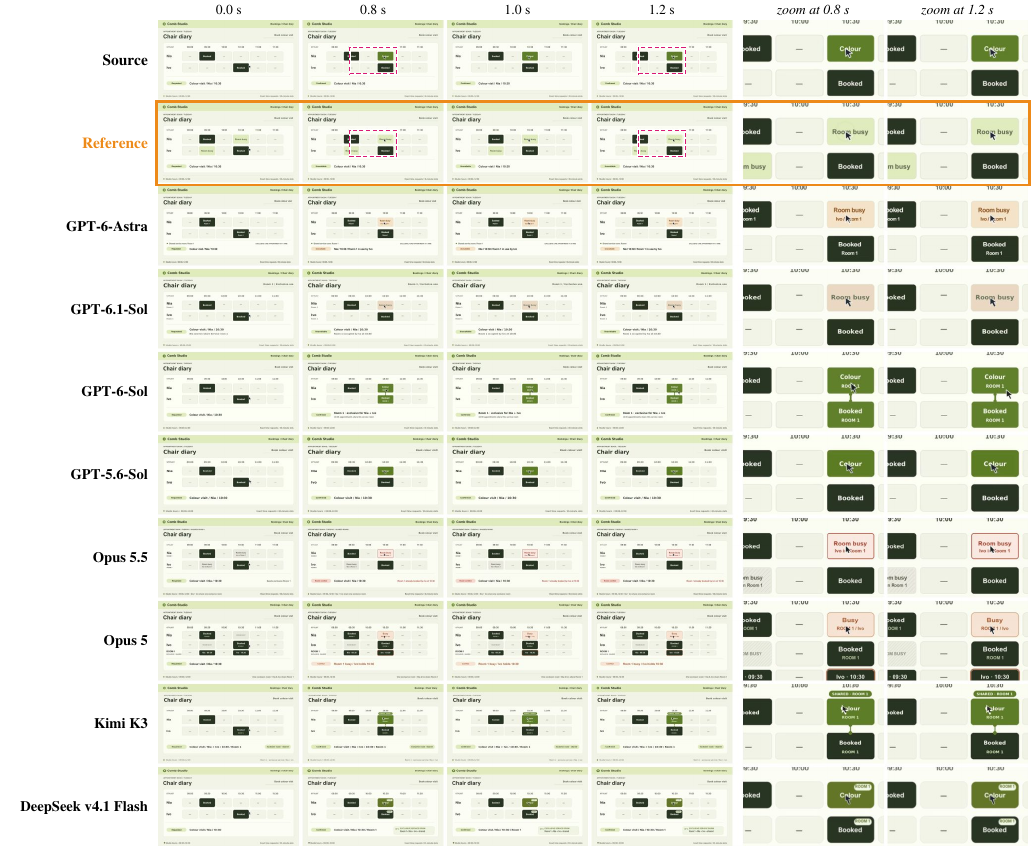}
\caption{\gallerycaptionfont \textbf{Web interfaces / Salon calendar / L3.} \emph{Instruction:} ``Make the bookings of Nia and Ivo share a single exclusive service room.'' \textbf{Direct:} (D1)~At the start of the video, when Nia already has a 09:30 booking, does Ivo's 09:30 cell also indicate that the room cannot be booked? (D2)~At the start of the video, when Ivo already has a 10:30 booking, does Nia's 10:30 cell also indicate that the room cannot be booked? \textbf{Derived:} (R1)~After the booking attempt, does Nia's 10:30 cell still show the room as occupied? (R2)~At the end of the video, below the schedule, is the request status beside Colour visit / Nia / 10:30 Unavailable? \textbf{Invariant:} (I1)~The two staff members, the time-slot grid and the interaction of sending a booking request are kept.}
\label{fig:g-web-15-salon-calendar-L3}
\end{figure*}

\begin{figure*}[p]
\centering
\pdfbookmark[2]{Web interfaces / Workflow builder / L3}{bm-web__17-workflow__L3}
\includegraphics[width=\linewidth]{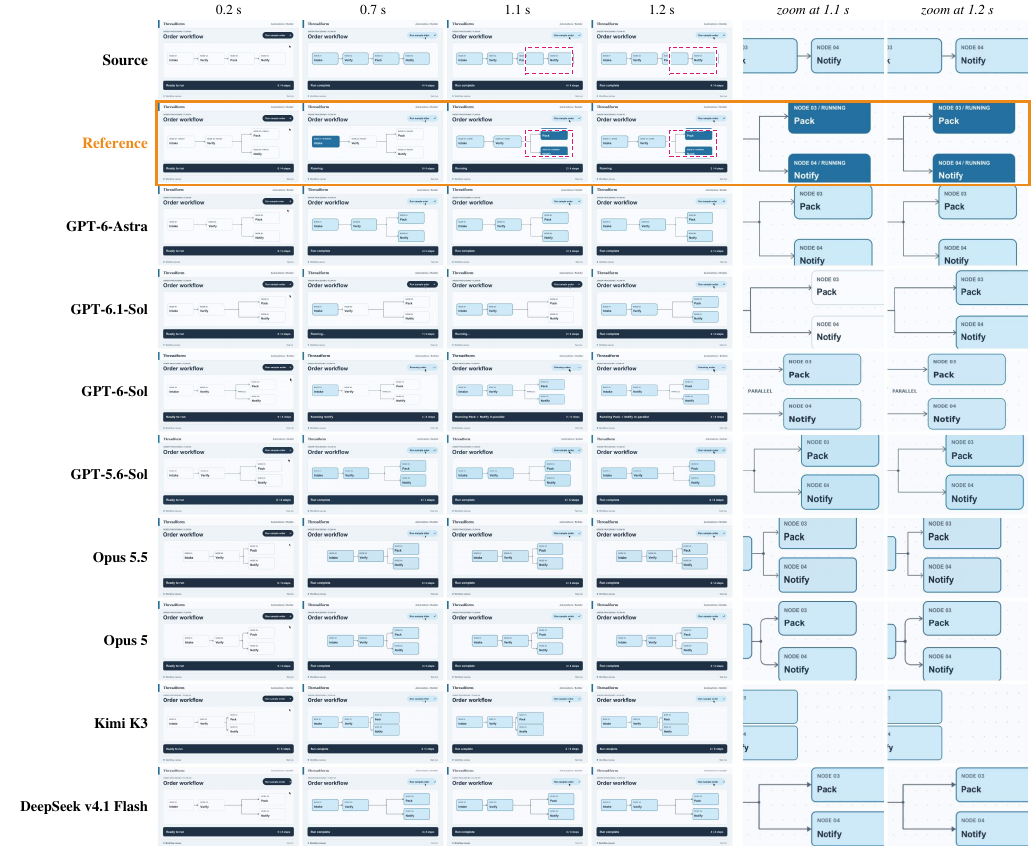}
\caption{\gallerycaptionfont \textbf{Web interfaces / Workflow builder / L3.} \emph{Instruction:} ``Make Pack and Notify run in parallel after Verify.'' \textbf{Direct:} (D1)~In the flow diagram, does the output of Verify connect to Pack and to Notify separately? (D2)~In the flow diagram, is there no serial arrow between Pack and Notify? \textbf{Derived:} (R1)~While Verify is shown as running, are Pack and Notify both still waiting? (R2)~After Verify is shown as done, is there a stretch of frames in which Pack and Notify are both marked as running at the same time? (R3)~At the end of the video, is the run count 4 / 4 steps? \textbf{Invariant:} (I1)~The Intake starting point and the demonstration of the workflow's states from waiting through running to done are kept.}
\label{fig:g-web-17-workflow-L3}
\end{figure*}

\clearpage

\fi

\end{document}